\documentclass[ijds,nonblindrev]{informs-ijds}

\OneAndAHalfSpacedXI

\usepackage{algorithm}
\usepackage{booktabs}
\usepackage{makecell}

\usepackage{algorithmic}
\usepackage{natbib}
 \bibpunct[, ]{(}{)}{,}{a}{}{,}%
 \def\bibfont{\small}%
 \def\Beta{\textrm{Beta}}
 \def\x{\mathbf{x}}

\def\NKT{N_{K_T}}
\def \NBT{N_{B_T}}
\def \NB{N_{B}}
\usepackage{amssymb,amsmath, bm, color}
\usepackage{graphicx}
\usepackage{natbib}
\usepackage{authblk}
\usepackage{hyperref}
\hypersetup{
    colorlinks=true,
    linkcolor=blue,
    filecolor=magenta,      
    urlcolor=cyan,
}
\usepackage{caption}
\usepackage{subcaption}
\usepackage{xcolor}

\newcommand{\RN}[1]{%
  \textup{\uppercase\expandafter{\romannumeral#1}}%
}

\TheoremsNumberedThrough     
\ECRepeatTheorems

\EquationsNumberedThrough    

\begin{document}



\RUNTITLE{Sparse Density Trees and Lists}

\TITLE{
Sparse Density Trees and Lists: An Interpretable Alternative to High-Dimensional Histograms$^1$
\def\thefootnote{1}\footnotetext{
Our code is available at \url{https://github.com/shangtai/githubcascadedmodel}.\newline
Data Ethics Note: There are no ethical issues with this algorithm that we are aware of. Data sets for testing the algorithm are either simulated or publicly available through the UCI Machine Learning Repository. The housebreak data was obtained through the Cambridge Police Department, Cambridge MA. 
\def\thefootnote{\arabic{footnote}}
}}



\ARTICLEAUTHORS{%
\AUTHOR{Siong Thye Goh$^*$}
\AFF{Singapore Management University, Singapore, 188065, \EMAIL{stgoh@smu.edu.sg}} 
\AUTHOR{Lesia Semenova$^*$}
\def\thefootnote{*}\footnotetext{Authors contributed equally.}\def\thefootnote{\arabic{footnote}}
\AFF{Department of Computer Science, Duke University, Durham, NC, 27708, \EMAIL{lesia@cs.duke.edu}}
\AUTHOR{Cynthia Rudin}
\AFF{Department of Computer Science, Duke University, Durham, NC, 27708, \EMAIL{cynthia@cs.duke.edu}}
}

\ABSTRACT{%
We present sparse tree-based and list-based density estimation methods for binary/categorical data. Our density estimation models are higher dimensional analogies to variable bin width histograms. In each leaf of the tree (or list), the density is constant, similar to the flat density within the bin of a histogram. Histograms, however, cannot easily be visualized in more than two dimensions, whereas our models can. The accuracy of histograms fades as dimensions increase, whereas our models have priors that help with generalization. Our models are sparse, unlike high-dimensional fixed-bin histograms. We present three generative modeling methods, where the first one allows the user to specify the preferred number of leaves in the tree within a Bayesian prior. The second method allows the user to specify the preferred number of branches within the prior. The third method returns density lists (rather than trees) and allows the user to specify the preferred number of rules and the length of rules within the prior. The new approaches often yield a better balance between sparsity and accuracy of density estimates than other methods for this task. We present an application to crime analysis, where we estimate how unusual each type of modus operandi is for a house break-in.
}%



\maketitle

\section{Introduction}

A histogram is a piecewise constant density estimation model. 
There are good reasons that the histogram is among the first techniques taught to any student dealing with data \citep{chakrabarti2006data}: (i) histograms are easy to visualize, (ii) they are accurate as long as there are enough data in each bin, and (iii) they have a logical structure that most people find interpretable. A downside of the conventional histogram is that all of these properties fail in more than two or three dimensions, particularly for binned binary or categorical data. One cannot easily visualize a conventional higher dimensional bar plot or histogram. For binary data, this would require us to visualize bins on a high dimensional hypercube. Worse, there may not be enough data in each bin, so the estimates would cease to be accurate. In terms of interpretability, for a higher dimensional histogram, a large set of logical conditions characterizing each bin ceases to be an interpretable representation of the data, and can easily obscure important relationships between variables. Considering all of the marginals is often useless for binary variables, since there are only two bins (0 and 1). Not only do histograms become uninterpretable in high dimensions, other high-dimensional density estimation methods are also uninterpretable: flexible nonparametric approaches such as kernel density estimation simply produce a formula, and the estimated density landscape cannot be easily visualized without projecting it to one or two dimensions, in which case we would lose substantial information. The question we ask is how to construct a piecewise constant density estimation model (like a histogram) for categorical data that has the three properties mentioned above: (i) it can be visualized, (ii) it is accurate, (iii) it is interpretable.

In this paper, we present three methods for constructing tree- and list-based density estimation models. These types of models are alternatives to bar plots or variable bin-width histograms \citep[e.g., see][]{wand1997data,scott1979optimal}. In our models, a leaf is analogous to a histogram bin (i.e., a probability mass function bin) and is defined by conditions on a subset of variables (e.g., if $\x$ is the feature vector, the conditions can be ``the second component of $\x$ is 0" and ``the first component of $\x$ is 1''), and the density is estimated to be constant with each leaf (that is, regardless of what the other components of $\x$ are, the density is constant). Our approaches use only a subset of the variables to define the bins, making them more interpretable. 

\begin{figure}[ht]
\centering
\includegraphics[width=\columnwidth]{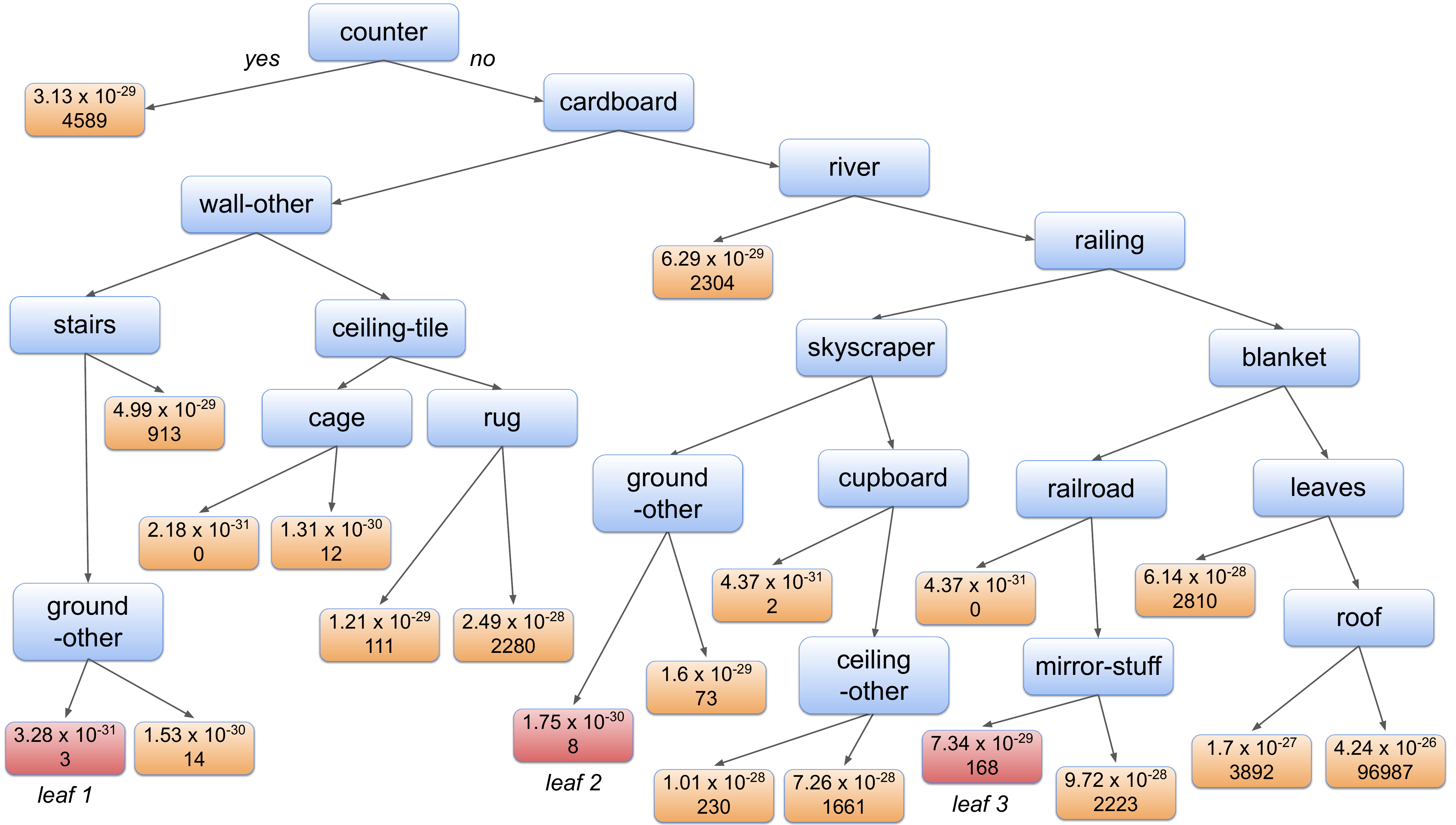}
\caption{A sparse density tree to represent the COCO-stuff labels. Each leaf (orange or red color) shows the density and number of training points that belong to that leaf (since densities are small for large datasets such as COCO-stuff labels, the number of points might be easier to understand). Leaves 1, 2, and 3 are visualized in Figure \ref{fig:coco_leaves}. This tree contains 20 leaves. It took approximately 25 seconds to create the tree and around 6.4 minutes to run the validation process that tunes parameters and optimizes the tree for each parameter setting. Each run takes approximately 25 seconds; there are 5 repeats per parameter setting, and 3 parameter settings.}
\label{fig:example_coco}
\end{figure}

\begin{figure*}[t]
\captionsetup[subfigure]{justification=centering, font=small}
	\centering
	\begin{subfigure}[t]{0.35\textwidth}
		\centering
		\includegraphics[height=1.5in]{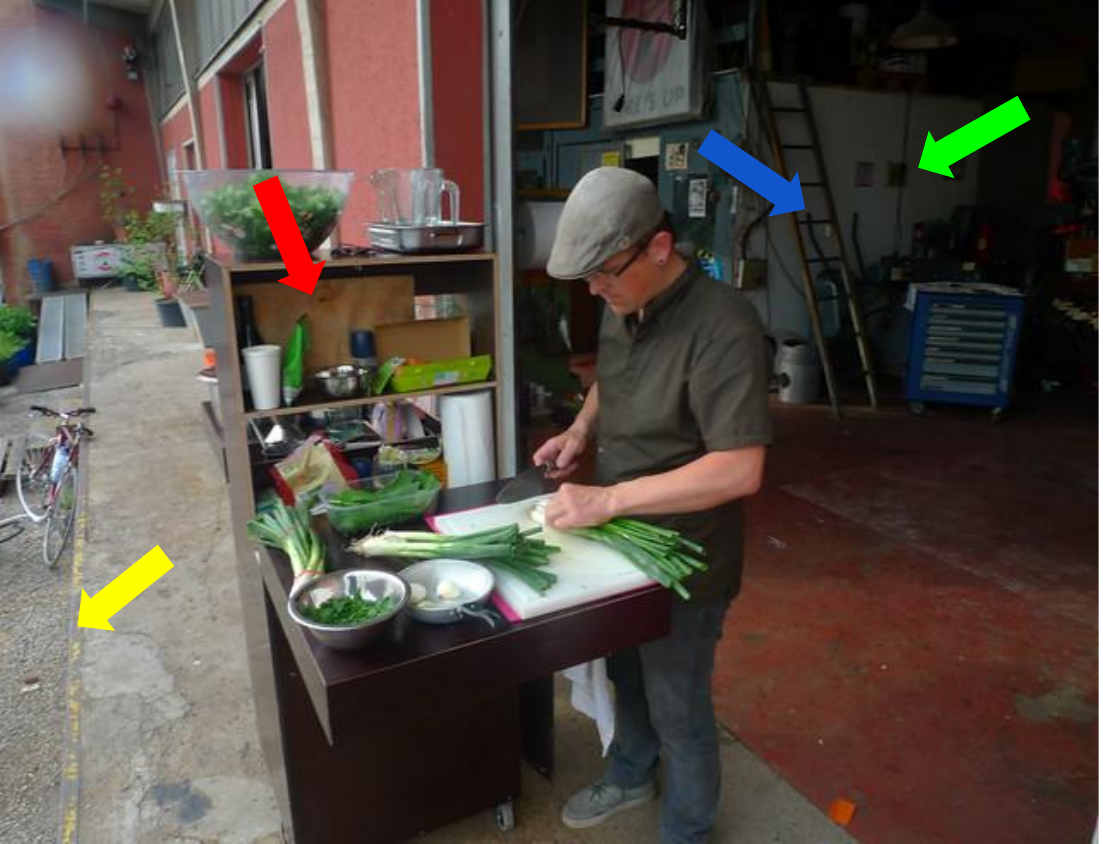}
		\caption{\textcolor{red}{red} -- \textit{cardboard}\\
\textcolor{green}{green} -- \textit{wall-other}\\
\textcolor{blue}{blue} -- \textit{stairs}\\
\textcolor{yellow}{yellow} -- \textit{ground-other}}		
	\end{subfigure}
	\hspace{0.5cm}
	\begin{subfigure}[t]{0.35\textwidth}
		\centering
		\includegraphics[height=1.5in]{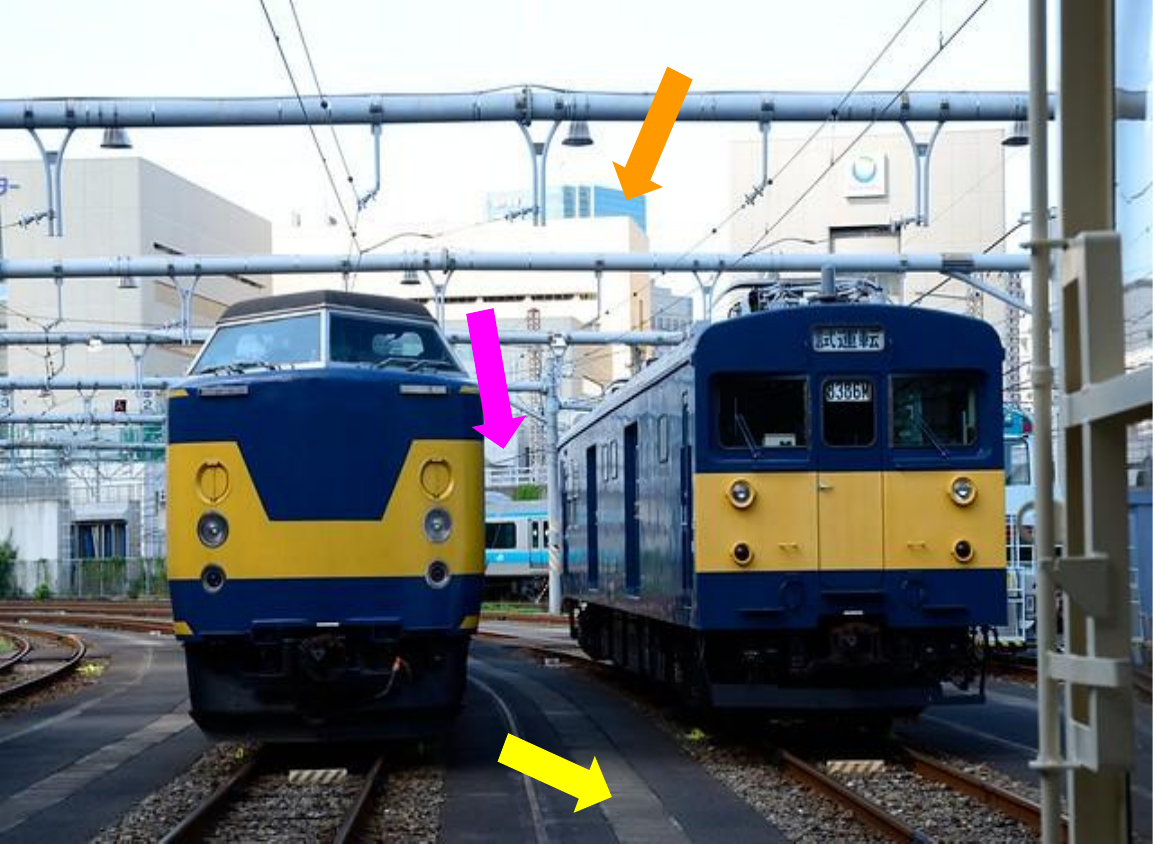}
		\caption{\textcolor{magenta}{magenta}  -- \textit{railing} \\
\textcolor{orange}{orange} -- \textit{skyscraper}\\
\textcolor{yellow}{yellow} -- \textit{ground-other}}
	\end{subfigure}
	\hspace{0.5cm}
	\begin{subfigure}[t]{0.2\textwidth}
		\centering
		\includegraphics[height=1.5in]{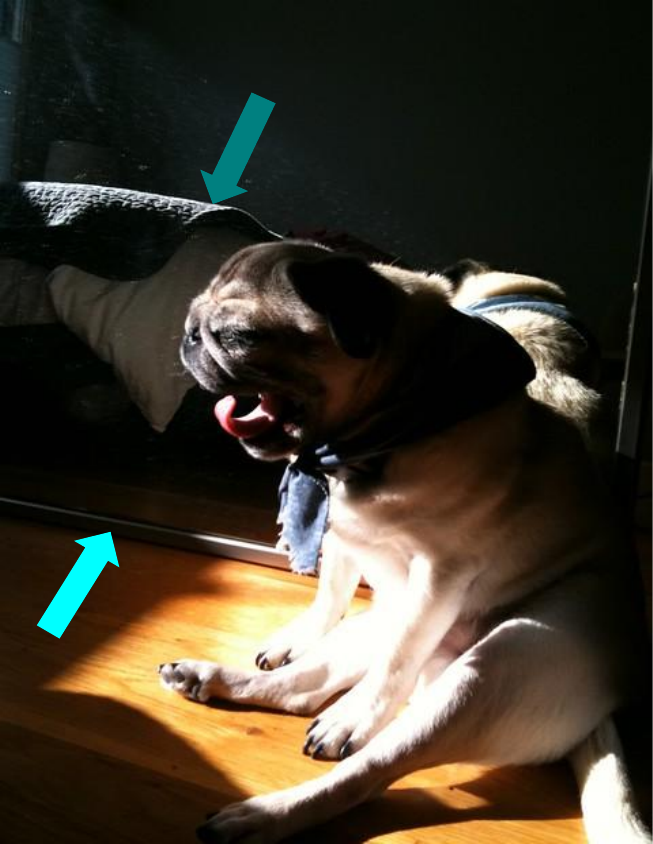}
		\caption{\textcolor{teal}{teal} -- \textit{blanket}\\
\textcolor{cyan}{aqua} -- \textit{mirror-stuff}}
	\end{subfigure}
	\caption{Examples of images that contain labels from leaf 1 (a), 2 (b) and 3 (c) in Figure \ref{fig:example_coco}. 
	}\label{fig:coco_leaves}
\end{figure*}

Our density estimation models can be useful in multiple domains to detect new patterns or errors in the data. For example, Figure \ref{fig:example_coco} shows the sparse density tree for the COCO-Stuff \citep{caesar2018cvpr} labels. The data set contains 118k training samples over 91 stuff categories. While the labels are sparse, our method finds interesting combinations, such as \textit{mirror} and \textit{blanket}, or \textit{railing}, \textit{skyscraper}, and \textit{ground} that are shown in Figure \ref{fig:coco_leaves}. 

\begin{figure}[ht]
\centering
\includegraphics[width=0.9\columnwidth]{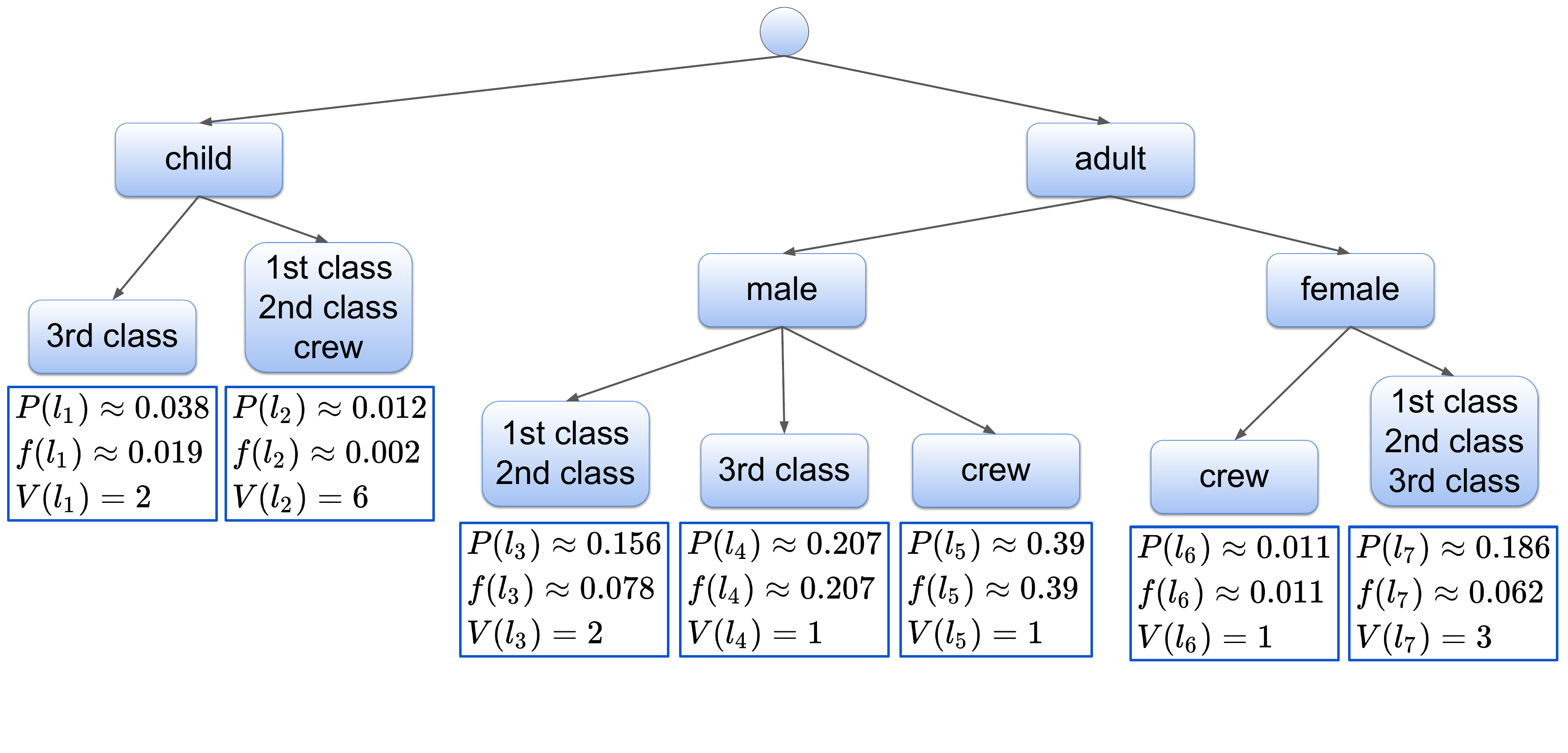}
\caption{A sparse density tree to represent the Titanic data set. Probability of belonging to the leaf ($P$), the densities ($f$) and volume ($V$) are specified for each leaf of the sparse tree. The density is estimated to be constant within each leaf. Here, we can see that the volume times the density equals the probability to be in the leaf. We formally define probability, density, and volume in Section \ref{sec:methods}.\label{fig:example_titanic}}
\end{figure}

Let us give a second example, this time to illustrate how each bin is modeled to be of constant density. We are interested in understanding the set of passengers on the Titanic. Specifically, our goal is to create a density model to understand how common or unusual the particular details of a passenger might be, given three categorical features: passenger class (1st, 2nd, 3rd, crew), whether someone is an adult (adult, child) and gender (male, female). A leaf (histogram bin) in our model might be: if \textit{passenger class} is \textit{3rd class} and a person is \textit{child}, then the probability of belonging to this leaf is $p(\textrm{state})$ is 0.038. That is, the total density in the bin where these conditions hold is 0.038, so 3.8\% of passengers are children traveling in 3rd class. The density is constant within the leaf, so for an additional variable \textit{gender} not in the tree, with values \texttt{male}, \texttt{female}, each of these values would be equally probable in the leaf, each with probability $0.038/2=0.019$. 
We described just one bin above, whereas a full tree is in Figure \ref{fig:example_titanic}.

Each of our methods aim to globally optimize a Bayesian posterior possessing a sparsity prior, which acts as a regularization term.  \citep[Unlike past work on density trees,][ our three methods are not constructed using greedy tree induction, they are optimized instead.]{Ram:2011:DET:2020408.2020507}
In this way, Bayesian priors control the shape of the tree or list. This helps with both generalization and interpretability. For the first of our three methods, the prior parameter controls the number of leaves; its value should be set to be the user's desired number of leaves. For the second method, the prior controls the desired number of branches for nodes of the tree. For the third method, which creates lists (one-sided trees), the prior controls the desired number of leaves and also the length of the leaf descriptions. 

The structure of these sparse tree models aims to fix the three issues with conventional histograms: (i) visualization: we need only write down the conditions we used in the density tree (or density rule list) to visualize the model. 
 (ii) accuracy: the prior encourages the density estimation model to be smaller, which means the bins are larger, and generalize better.  (iii) interpretability: the prior encourages sparsity, and encourages the model to obey a user-defined notion of interpretability. 
 
 In what follows, we provide related works and then we describe our three methods in Section \ref{sec:methods}. In Section \ref{sec:optimization} we discuss how we optimized the posteriors for three methods. Section \ref{sec:experiments} provides experiments and examples, Section \ref{sec:perf_analysis} provides a run-time analysis, and a study using simulated data sets. Section \ref{sec:consistensy} provides a consistency proof, and we conclude in Section \ref{sec:conclusion}.

\section{Related Work}

\textit{Nonparametric density estimation.} Density estimation is a classic topic in statistics and unsupervised machine learning.
Without using domain-specific generative assumptions, the most useful techniques have been nonparametric, mainly variants of kernel density estimation (KDE) \citep{akaike1954approximation,rosenblatt1956remarks,parzen1962estimation,cacoullos1966estimation,mahapatruni2011cake,nadaraya1970remarks,rejto1973density,wasserman2006all,silverman1986density,devroye1991exponential,cattaneo2019simple,varet2019numerical}. KDE is highly tunable, not domain dependent, and can generalize well, but does not have the interpretable logical structure of histograms. Similar alternatives include mixtures of Gaussians \citep{NIPS1999_1673,zhuang1996gaussian,ormoneit1995improved,ormoneit1998averaging,chen2006probability,seidl2009indexing}, forest density estimation \citep{Liu:2011:FDE:1953048.2021032}, RODEO \citep{AISTATS07_LiuLW} and other nonparametric Bayesian methods \citep{muller2004nonparametric}
 which have been proposed for general purpose (not interpretable per se) density estimation. \citet{jebara2012bayesian} provides a Bayesian treatment of latent directed graph structure for non-iid data, but does not focus on sparsity. P\'olya trees have been generated probabilistically for real valued features \citep{wong2010optional} and could be used as priors for our method. \citet{friedman1984projection} uses a projection pursuit method to perform density estimation. Another task related to density estimation is level set estimation, where the goal is to determine whether the density at a leaf is higher than a prespecified value $\gamma$; \citet{willett2007minimax} address the problem using tree representations, and \citet{holmstrom2015estimation} use a discretized kernel to construct level set trees. Some works \citep[e.g., ][]{sasaki2018neural, Liue2101344118} use neural networks to perform density estimation, which do not aim to be interpretable. 
In \cite{luo2019combining}, a smoothing spline is used to perform density estimation. In \cite{rehn2018forest}, a non-parametric density estimator called ``FRONT'' segments a data stream through a periodically updated linear transformation.

 The most closely related paper to ours is on density estimation trees (DET) \citep{Ram:2011:DET:2020408.2020507} and its extensions.  DETs are constructed in a top-down greedy way. This gives them a disadvantage in optimization, often leading to lower quality trees. They also do not have a generative interpretation, and their parameters do not have a physical meaning in terms of the shape of the trees (unlike the methods defined in this work). DET was used by \citet{wu2018density}, leveraging ideas from \citet{lu2013multivariate} with random forests to perform density estimation. DET has also been applied to high energy physics \citep{anderlini2015density}. Techniques to avoid overfitting in tree-based density estimation models have been discussed by \citet{anderlini2016density}. Other top-down greedy approaches \citep[e.g.,][]{yang2014density,yang2014discovering,yang2015density} use discrepancy, negative log-likelihood, or MISE \citep{ooi2012density} as splitting criteria. A distinction between our work and existing work is that we place priors \textit{directly} on the shape of the trees that we desire, using a Bayesian approach. We do not rely on greedy splitting and pruning, the splits are optimized instead.

\textit{Bayesian Tree Models}
Bayesian tree models are commonly used for tasks other than density estimation (i.e., classification and regression). Some examples include Bayesian CART \citep{wu2007bayesian}, Bayesian Additive Regression Trees (BART) \citep{chipman2010bart}, and Bayesian Rule Lists \citep{LethamRuMcMa15,YangRuSe16}. Bayesian CART and BART use priors that specify the probability that a node is terminal and a uniform probability distribution over the choices for a split. Our priors function differently. Often, we have priors over \textit{global} properties of the trees such as the number of total leaves (our Method I and Method III). Also, we have prior parameters governing the number of branches at a node (our Method II), which is different from Bayesian CART or BART where there are only two branches at every node. Our rule list density approach (Method III) has a prior on the number of conditions used in each rule, which is similar to Bayesian Rule Lists, but not similar to Bayesian CART or BART, which have only one condition defining each split.


\section{Methods}\label{sec:methods}

We use a Bayesian approach to achieve sparsity, by introducing priors on the shape of the trees. In particular, in Method 1, we define a prior on the number of leaves in the tree, then calculate the likelihood of the data having been generated by a particular tree, and multiply the prior and the likelihood to create a posterior to be optimized over all trees. In Method 2, we instead choose a prior over the number of branches for each split in the tree, preferring a small number of branches. In Method 3, we switch to rule lists, where the prior prefers models with a smaller number of rules and a smaller number of conjunctions per rule.


Before the introduction of the three methods, we first present notation. We will focus on problem of estimating the unknown distribution $f$ with tree-structured approximations given a set of $n$ data points $X = \{x_i,...,x_n\}$ drawn i.i.d. from $f$ on $\mathcal{X}\subset \mathbb{R}^p$. 

\begin{figure}[ht]
\centering
\includegraphics[width=6cm]{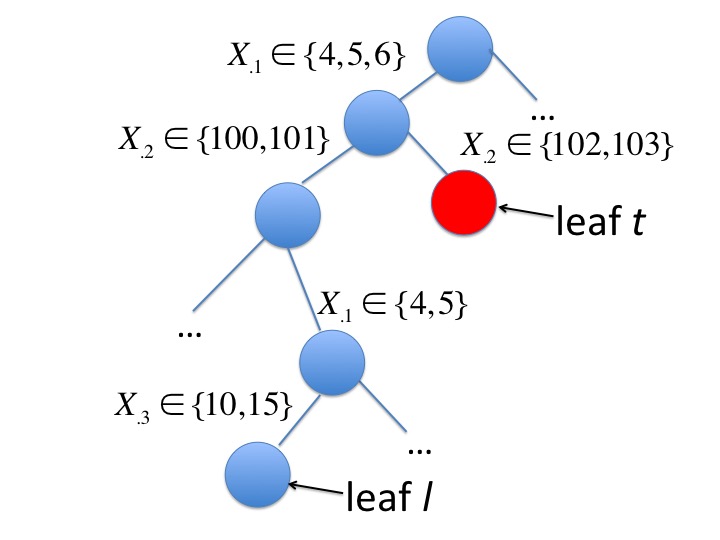}
\caption{Example of computation of volume.\label{MyTree}}
\end{figure}

For the tree-structured estimations, we express the path to a leaf as the set of conditions on each of $p$ features along the path. For instance, for a particular leaf (\textcolor{red}{ leaf $t$ in Figure \ref{MyTree}}), we might see conditions on the first feature, e.g., $x_{.1} \in \left\{ 4,5,6\right\}$ and the second feature $x_{.2} \in \left\{ 100,101\right\}$. Thus the leaf is defined by the set of all feature values that obey these conditions, that is, the leaf could be $$x \in\left\{ x_{.1} \in \left\{4,5,6 \right\}, x_{.2} \in \left\{ 100, 101\right\},  x_{.3}, x_{.4}, \ldots, x_{.p} \textrm{ any values}\right\}.$$ 
This implies there is no restriction on $ x_{.3}, x_{.4}, \ldots, x_{.p}$ for observations within the leaf.
Notationally, a condition on the $j^{th}$ feature is denoted $x_{.j} \in \sigma_j(l)$ where $\sigma_j(l)$ is the set of allowed values for feature $j$ along the path to leaf $l$. If there are no conditions on feature $j$ along the path to $l$, then $\sigma_j(l)$ includes all possible values for feature $j$. 
Thus, leaf $l$ includes feature values $x$ obeying:
$$x \in \left\{ x_{.1} \in \sigma_1(l),x_{.2} \in \sigma_2(l), \ldots, x_{.p} \in \sigma_p(l)\right\}.$$ 
For categorical data, the volume in a leaf $l$, which is required for computing the density, is calculated as:
\begin{equation}\label{eq:volume}
\mathbb{V}_l=\prod_{j=1}^p |\sigma_j(l)|.
\end{equation}
 We give an example of this computation next.

\noindent\textbf{Volume Computation Example.} 
The data are categorical. Possible values for $x_{.1}$ are $\left\{ 1,2,3,4,5,6,7\right\}$. Possible values for $x_{.2}$ are $\left\{ 100,101,102,103\right\}$. Possible values for $x_{.3}$ are $\left\{ 10,11,12,13,14,15\right\}$, and for $x_{.4}$ are $\left\{ 8,9,10\right\}$.
Consider the tree in Figure \ref{MyTree}. 

We compute the volume for leaf $l$. Here,
$\sigma_1(l)=\left\{ 4,5 \right\}$ since $l$ requires both $x_{.1} \in \left\{ 4,5,6\right\}$ and $x_{.1} \in \left\{ 4,5\right\}$. $\sigma_2(l)=\left\{ 100,101\right\},$ $\sigma_3(l)=\left\{ 10,15\right\},$ and $\sigma_4(l)=\left\{ 8,9,10\right\}$ because there is no restriction on $x_{.4}$. So $\mathbb{V}_l=\prod_j |\sigma_j(l)|=2\cdot 2\cdot 2\cdot 3=24.$



Our notation handles only categorical data (for ease of exposition) but can be extended to handle ordinal and continuous data. For ordinal data, the definition is the same as for categorical but $\sigma_j$ can (optionally) include only contiguous values (e.g. $\left\{ 3,4,5\right\}$ but not $\left\{ 3,4,6\right\}$). For continuous variables, $\sigma_j$ is the ``volume'' of the continuous variables, for example, for node condition $x_{.j} \in (0,0.5), \sigma_j=0.5-0$. 



In the next subsections, we present the three methods: the leaf-sparse modeling approach, the branch-sparse modeling approach, and a density list approach. We define the prior and likelihood for each of the three modeling methods, combining them to get posteriors.

\subsection{Method \texorpdfstring{\RN{1}}@{:} Leaf-Sparse Density Trees}

Let us define the prior, likelihood, and posterior for Method 1.


\noindent \textbf{Prior:} For this modeling method, the main prior on tree $T$ is on the number of leaves $K_T$.  This prior we choose to be Poisson with a particular scaling (which will make sense later on), where the Poisson is centered at a user-defined parameter $\lambda$, which is the user's desired number of leaves in the tree prior to seeing data. Notation $\NKT$ is the number of trees with $K_T$ leaves. It can be calculated directly given $K_T$. The prior over trees is: 
\begin{eqnarray*}
P(K_T | \lambda) = P(K_T \textrm{ is the number of leaves in tree }T| \lambda) &\propto& \NKT\cdot \textrm{ Poisson}(K_T,\lambda) \\
&=& \NKT e^{-\lambda}\frac{\lambda^{K_T}}{K_T!}.  
\end{eqnarray*}
Thus, $\lambda$ allows the user to control the number of leaves in the tree. The number of possible trees is finite, thus the distribution can be trivially normalized.
Among trees with $K_T$ leaves, tree $T$ is chosen uniformly, with probability $1/\NKT$. This means the probability to choose a particular tree $T$ is Poisson:
\begin{eqnarray*}
P(T|\lambda)\propto 
P(T|K_T)P(K_T|\lambda)&\propto& \frac{1}{\NKT}\NKT e^{-\lambda}\frac{\lambda^{K_T}}{K_T!} = e^{-\lambda}\frac{\lambda^{K_T}}{K_T!}\\ 
&\propto&
\textrm{Poisson}(K_T,\lambda).
\end{eqnarray*}

We place a uniform prior over the probabilities for a data point to land in each of the leaves. To do this, we start from a Dirichlet distribution with equal parameters $\alpha_1=\ldots=\alpha_{K_T}=\alpha \in \mathbb{Z}^+$ where hyperparameter $\alpha>1$. $\alpha$ is a pseudocount that is typically chosen to be a small number (1 or 2) to avoid a 0 value for the estimated densities. We denote the vector with $K_T$ equal entries $[\alpha,...,\alpha]$ as $\bm{\alpha}_{K_T}$. 
We draw multinomial parameters $\bm\theta=[{\theta}_1, \ldots, {\theta}_{K_T}]$ from Dir$(\bm{\alpha}_{K_T})$, which govern the prior on popularity of each leaf. 
Thus, the prior on feature values, given the tree structure is: $$p({\bm{\theta}}|\bm{\alpha}_{K_T},T)=\frac{1}{\Beta(\bm{\alpha}_{K_T})}\prod_{l=1}^{K_T} {\theta}_l^{{\alpha}-1}.$$ 
Here, $\Beta(\bm{\alpha}_{K_T})=\frac{\prod_{l=1}^{K_T} \Gamma(\alpha_{K_T,l})}{\Gamma(\sum_{l=1}^{K_T} \alpha_{K_T,l})}=\frac{(\Gamma({\alpha}))^{K_T}}{\Gamma(K_T{\alpha})}$ 
is the multinomial beta function which is also the normalizing constant for the Dirichlet distribution.

Thus, the first part of our generative modeling method is as follows, given hyperparameters $\lambda$ and $\alpha$: Number of leaves in $T$: $K_T \propto  \textrm{scaled Poisson}(\lambda), \textrm{ i.e., } \NKT \times \textrm{ Poisson}(K_T,\lambda)$, the tree shape, $T$ $\propto \textrm{Uniform over trees with }K_T\textrm{ leaves}$, and $\textrm{Prior distribution over leaves: }  \bm\theta \propto 
\textrm{Dir}(\bm{\alpha}_{K_T}).$
As usual, the prior is regularization that can be overwhelmed with a large amount of data. 
 However, it takes a substantial amount of data and an enormous number of feature-value pairs to reduce the prior's influence on the posterior.

\noindent \textbf{Likelihood:} Here we calculate the likelihood of the data to have arisen from a particular tree.
Let $n_l$ denote the number of points captured by the $l$-th leaf, and denote $\mathbb{V}_l$ to be the volume of that leaf, defined in \eqref{eq:volume}. The probability to land at any specific value within leaf $l$ is $\frac{\theta_l}{\mathbb{V}_l}$. The likelihood for the full data set is thus $$P(X|\bm{\theta},T)=\prod_{l=1}^{K_T} \left( \frac{\theta_l}{\mathbb{V}_l} \right)^{n_l}.$$\\
\textbf{Posterior:} 
The posterior, which is (by definition) the likelihood times the prior, can be written as follows, where we have substituted the distributions from the prior into the formulas. 
\begin{align*}
P(T|\lambda, \bm{\alpha},X) 
& \propto \int_{\bm{\theta}: \textrm{simplex}} P(K_T|\lambda)\cdot P(T|K_T)\cdot P(\bm{\theta}|\bm{\alpha}_{K_T},T)\cdot P(X|\bm{\theta},T) d \bm{\theta}\\
& \;\;\textrm{i.e.,} \int_{\bm{\theta}: \textrm{simplex}} \textrm{(size)} \cdot \textrm{(tree shape $|$ size)} \cdot \textrm{(leaf popularity)} \cdot \\
&\hspace*{30pt}\textrm{(data $|$ tree shape, leaf popularity) } d \bm{\theta}\\
&\propto\int_{\bm{\theta}: \textrm{simplex}} P(T|\lambda) \left[\frac{1}{\Beta(\bm{\alpha}_{K_T})} \left(\prod_{l=1}^{K_T} {\theta}_l^{\alpha-1} \right)\right]\left[\prod_{l=1}^{K_T} \left(\frac{{\theta}_l}{\mathbb{V}_l} \right)^{n_l} \right] d \bm{\theta}\\
&\propto P(T|\lambda) \frac{1}{\Beta(\bm{\alpha}_{K_T})} \left( \prod_{l=1}^{K_T}\left(\frac{1}{\mathbb{V}_l} \right)^{n_l}\right)\int_{\bm{\theta}: \textrm{simplex}}  \prod_{l=1}^{K_T} {\theta}^{n_l+\alpha-1}    d \bm{\theta} \\
&\propto P(T|\lambda)\frac{\Beta([n_1+\alpha, \ldots, n_{K_T}+\alpha])}{\Beta(\bm{\alpha}_{K_T})}\prod_{l=1}^{K_T}\frac{1}{\mathbb{V}_l^{n_l}} \\
&\propto P(T|\lambda) \frac{\Gamma (K_T \alpha)}{\Gamma (n+K_T\alpha)}\prod_{l=1}^{K_T} \frac{(n_l+\alpha-1)!}{(\alpha-1)!} {\mathbb{V}_l^{-n_l}},
\end{align*}
where $P(T|\lambda)$ is simply Poisson($K_T,\lambda$) as discussed earlier.
For numerical stability, we maximize the log-posterior which is equivalent to maximizing the posterior. 

As compared to full Bayesian approaches, by maximizing the posterior, we leverage relatively faster computation time and optimize for a single model, which can be important for interpretability. However, in turn, we miss out on uncertainty information, which we could get by modeling the posterior density.

For the purposes of prediction, we are required to estimate the density that is being assigned to leaf $l$. This is calibrated to the data, simply as: $\hat{f}=n_l \slash (n \mathbb{V}_l)$ where $n$ is the total number of training data points and $n_l$ is the number of training data points that reside in leaf $l$. The formula implicitly states that the density in the leaf is uniformly distributed over the features whose values are undetermined within the leaf (features for which $\sigma_j$ contains all possible values for feature $j$). 
That is, the probability mass function is the same for any point within the leaf.

\subsection{Method \texorpdfstring{\RN{2}}@{:} Branch-Sparse Density Trees}





In Method I, we regularized the number of leaves of the density tree. In Method II, we instead regularize the number of branches at each node to simplify the model, by avoiding models with too many branches. In other words, Method I aims to have a small number of leaves. Method II aims to have a small number of branches.

In Method I, a Dirichlet distribution is drawn only over the leaves. In Method II, a Dirichlet distribution is drawn at every internal node to determine branching. Similar to the previous method, we choose the tree that optimizes the posterior.

\noindent \textbf{Prior:}
The prior is comprised of two pieces: the part that creates the tree structure, and the part that determines how data propagates through it.\\
\textbf{Tree Structure Prior}:
For tree $T$, we let $B_T= \left\{ b_i |i \in I \right\}$ be a multiset, where each element is the count of branches from a node of the tree. For instance, if in tree $T$ with three nodes, the three nodes have 3 branches, 2 branches, and 2 branches respectively, then $B_T=\left\{3,2,2\right\}$. We let $\NBT$ denote the number of trees with the same multiset $B_T$. Note that $B_T$ is unordered, so 
$\{3,2,2\}$ is the same multiset as $\{2,3,2\}$ or 
$\{2,2,3\}$.

Let $I$ denote the set of internal nodes of tree $T$ and let $L$ denote the set of leaves. As before, we let $\mathbb{V}_l$ denote the volume of leaf $l$. 

In the generative model for the Branch-sparse density tree, a Poisson distribution with parameter $\lambda$  is used at each internal node in a top down fashion to determine the number of branches. 
Iteratively, for node $i$, the number of branches, $b_i$, obeys $b_i \sim \textrm{Poisson}(\lambda),$ where the parameter $\lambda$ is the desired number of branches (before seeing data). Hence, at any node $i$, with probability $\exp(-\lambda) \frac{\lambda^{b_i}}{b_i!}$, there are $b_i$ branches from node $i$. This implies that with probability $\exp(-\lambda)$, the node is a leaf. In summary,
\begin{equation*} P(\text{Multiset of branches }= B|\lambda) \propto \NB \left[\prod_{i \in I}e^{-\lambda} \frac{\lambda^{b_i}}{b_i!}\right]\left[\prod_{l \in L}e^{-\lambda}\right].\end{equation*} 

Among trees with multiset $B$, tree $T$ is chosen uniformly, with probability $\frac{1}{\NB}.$ This means the probability to choose a particular tree shape is: 
\begin{align} P(T|\lambda)&\propto P(T|B_T)P(B_T|\lambda) \propto \frac{1}{\NBT} \NBT \left[\prod_{i \in I}e^{-\lambda} \frac{\lambda^{b_i}}{b_i!}\right]\left[\prod_{l \in L}e^{-\lambda}\right] \nonumber \\&=\left[\prod_{i \in I}e^{-\lambda} \frac{\lambda^{b_i}}{b_i!}\right]\left[\prod_{l \in L}e^{-\lambda}\right].  \label{branchtreeprior} \end{align}  

\noindent\textbf{Tree Propagation Prior:}
After the tree structure is determined, we need a generative process for how the data propagate through each internal node. We denote $\theta_{l}$ as the probability to land in leaf $l$. We denote $\widetilde{\theta}_{ij}$ as the probability to traverse to node $j$ from internal node $i$. Notation $\bm{\theta}$ is the vector of leaf probabilities (the $\theta_l$'s), $\widetilde{\bm{\theta}}$ is the set of all $\widetilde\theta_{ij}$'s,
and $\widetilde{\bm{\theta}}_i$ is the set of all internal node transition probabilities from node $i$ (the $\widetilde{\theta}_{ij}$'s).

We compute $P(\widetilde{\bm{\theta}_i}|\alpha, T)$ for all internal nodes $i$ of tree $T$. As before, $\alpha$ is a pseudo-count to avoid 0-valued estimated densities. At each internal node, we draw a sample from a Dirichlet distribution with parameter $[\alpha,\ldots, \alpha]$ (of size equal to the number of branches $b_i$ of $i$) to determine the proportion of data, $\widetilde{\theta}_{i,j}$, that should go along the branch leading to each child node $j$ from the internal parent node $i$. 
Thus, $\widetilde{\bm\theta_i} \sim \textrm{Dir}(\bm\alpha)$ for each internal node $i$, that is:
$$P(\widetilde{\bm{\theta}}_i|\bm\alpha,T)=\frac{1}{\Beta_{b_i}(\bm{\alpha})}\prod_{j \in C_i} \widetilde{\theta}_{ij}^{\alpha-1},$$ where $\Beta_k(\bm{\alpha})$ is the normalizing constant for the Dirichlet distribution with parameter ${\alpha}$ and $k$ categories, and $C_i$ are the indices of the children of $i$. Thus,
\begin{equation}\label{model2priorsecondterm}
P(\widetilde{\bm{\theta}}|\bm\alpha,T)=\prod_i P(\widetilde{\bm{\theta}}_i|\bm\alpha,T) = \prod_i
\frac{1}{\Beta_{b_i}(\bm{\alpha})}\prod_{j \in C_i} \widetilde{\theta}_{ij}^{\alpha-1}.\end{equation}

Thus, the prior is 
$P(T|\lambda) \cdot P(\widetilde{\bm{\theta}}|\alpha,T)$, where $P(T|\lambda)$ is in (\ref{branchtreeprior}) and $P({\widetilde{\bm\theta}}|\alpha,T)$ is in (\ref{model2priorsecondterm}).

In summary, the prior of Method II is as follows, given hyperparameters $\lambda$ and $\alpha$:

\begin{itemize}
\item Multiset of branches, $B_T \propto  \NBT \left[\prod_{i \in I}e^{-\lambda} \frac{\lambda^{b_i}}{b_i!}\right]\left[\prod_{l \in L}e^{-\lambda}\right],$ 
\item Tree shape, $T\sim \textrm{Uniform over trees with branches } B_T,$
\item Prior distribution over each branch, $\widetilde{\bm\theta}_i \sim 
\textrm{Dir}(\alpha).$
\end{itemize}

\noindent The likelihood is the same as that for Method I.

\noindent \textbf{Posterior:} 
The posterior is proportional to the prior times the likelihood terms. Here we are integrating over the $\widetilde{\bm{\theta}}_i$ terms for each of the internal nodes $i$.
\begin{align*}
&P(T|\lambda, \alpha, X) \\
&\propto \int P(B_T|\lambda) \cdot P(T|B_T) \cdot P(\widetilde{\bm{\theta}}| \alpha,T) \cdot P(X| \widetilde{\bm{\theta}},T) d\widetilde{\bm{\theta}} \\
&\propto \left[\prod_{l \in L}  \left(\frac{ e^{-\lambda}}{\mathbb{V}_l^{n_l}}\right)\right]\left(\prod_{i \in I} e^{-\lambda} \frac{\lambda^{b_i}}{b_i!}\frac{1}{\Beta_{b_i}([\alpha, \ldots, \alpha])} \int_{\widetilde{\bm{\theta}}_i \in \text{simplex}}  \prod_{c \in C_i} \widetilde{\theta}_{i,c}^{\alpha-1} \widetilde{\theta}_{i,c}^{n_{c}}d \widetilde{\bm{\theta}}_i \right) \\
&= e^{-\lambda(|I|+|L|)}
\left(\prod_{i \in I}  \frac{\lambda^{b_i}}{b_i!}\frac{1}{\Beta_{b_i}([\alpha, \ldots, \alpha])} \int_{\widetilde{\bm{\theta}}_i \in \text{simplex}}  \prod_{c \in C_i} \widetilde{\theta}_{i,c}^{n_{c}+\alpha-1} d \widetilde{\bm{\theta}}_i \right) \prod_{l \in L}  \left(\frac{ 1}{\mathbb{V}_l^{n_l}}\right)\\
&= e^{-\lambda(|I|+|L|)} \lambda^{\sum_{i \in I} b_i}
\left(\prod_{i \in I}  \frac{1}{b_i!}\frac{\Beta_{b_i}([\alpha+n_{c_1}, \ldots, \alpha+n_{c_{b_i}}])}{\Beta_{b_i}([\alpha, \ldots, \alpha])}   \right) \prod_{l \in L}  \left(\frac{ 1}{\mathbb{V}_l^{n_l}}\right)\\
&= e^{-\lambda(|I|+|L|)} \lambda^{
|L|+|I|-1
}
\left(\prod_{i \in I}  \frac{1}{b_i!}\frac{B_{b_i}([\alpha+n_{c_1}, \ldots, \alpha+n_{c_{b_i}}])}{\Beta_{b_i}([\alpha, \ldots, \alpha])}   \right) \prod_{l \in L}  \left(\frac{ 1}{\mathbb{V}_l^{n_l}}\right)
\end{align*}
where $c_1,\ldots,c_{b_i} \in C_i$ in the second last expression.
We used the equation $\sum_{i \in I} b_i=|L|+|I|-1$ for a tree in the last line.

\noindent \textbf{Possible Extension:} We can include an upper layer of the hierarchical Bayesian model to control (regularize) the number of features $d$ (out of a total of $p$ dimensions) that are used in the model. This would introduce an extra multiplicative factor within the posterior of
 $\left( \begin{array}{c} p \\ d \end{array} \right)\gamma^d (1-\gamma)^{p-d}$, where $\gamma$ is a parameter between 0 and 1, where a smaller value favors a simpler model. $\gamma$ corresponds to the probability that a feature is chosen to be included in the model. For example, the value 0.5 corresponds to the case where the user prefers to have half of the features to be chosen. The posterior would become:
\begin{eqnarray*}
\lefteqn{\left( \begin{array}{c} p \\ d \end{array} \right)\gamma^d (1-\gamma)^{p-d} e^{-\lambda(|I|+|L|)} \lambda^{|I|+|L|-1}} \\
&&\left(\prod_{i \in I}  \frac{1}{b_i!}\frac{\Beta_{b_i}([\alpha+n_{c_1}, \ldots, \alpha+n_{c_{b_i}}])}{\Beta_{b_i}([\alpha, \ldots, \alpha])}   \right) \prod_{l \in L}  \left(\frac{ 1}{\mathbb{V}_l^{n_l}}\right).
\end{eqnarray*}

 \subsection{Method \texorpdfstring{\RN{3}}@{:} Sparse Density Rule List}

Rather than producing a general tree, an alternative approach is to produce a rule list, which is a one-sided tree. Rule lists are easier to optimize than trees. Each tree can be expressed as a rule list by creating a rule for each leaf, where the conditions defining the leaf also define the rule. By using lists, we implicitly hypothesize that the full space of trees may not be necessary and that simpler rule lists may suffice.

An example of a sparse density rule list is as follows:
\textbf{if} $x$ obeys $a_1$ \textbf{then} density$(x)=f_1$
\textbf{else if} $x$ obeys $a_2$ \textbf{then} density$(x)=f_2$
$\ldots$
\textbf{else if} $x$ obeys $a_m$ \textbf{then} density$(x)=f_m$
\textbf{else} density$(x)=f_0$.

Here, as with the trees, the density is the probability mass function, which is constant for the entire portion of the feature space that falls into the leaf.

The antecedents $a_1$,...,$a_m$ are Boolean assertions, that are either true or false for each data point $x_i$. They are chosen from a large pre-mined collection of possible antecedents, called $A$. 
We define $A$ to be the set of all possible antecedents of size at most $H$, where the user chooses $H$. The size of $A$ is:
$|A|=\sum_{j=0}^H A_j,$
where $A_j$ is the number of antecedents of size $j$,
$$A_j=\sum_{ \left[ \begin{array}{c} t_1, t_2, \ldots, t_j \in \left\{ 1,\ldots, p \right\} \\ \text{s.t. } t_1 > t_2 \ldots >t_j \end{array}\right]} \prod_{i=1}^j q_{t_i},$$
where feature $i$ consists of $q_i$ categories. 

For example, say the features consist of 2, 3, 4, and 5 categories respectively. If $H=2$, then the total number of elements of $A$ is $|A|=$ 1 (no feature is chosen) + 2 (because there are 2 categories for the first feature) + 3 (because there are 3 categories for the second feature) + 4 (because there are 4 categories from the third feature) + 4 (because there are 5 for the fourth feature) + 6 (possible combinations of feature 1 and feature 2) + $\ldots$ + 20 (possible combinations of feature 3 and feature 4). 

\noindent \textbf{Generative Process:} 
We now sketch the generative process for the tree from the observations $X = \{x_i\}$ and antecedents $A = \{a_j\}$. Prior parameter $\lambda$ is the user's preference of the length of the density list (in the absence of data), and $\eta$ is the user's preference for the number of conjunctions in each sub-rule $a_j$.

Define $a_{<j}$ as the antecedents before $j$ in the rule list if there are any. For example $a_{<3}=\left\{ a_1, a_2 \right\}$. Similarly, let $c_j$ be the cardinalities of the antecedents before $j$ in the rule list. Let $d$ denote the rule list. Following the exposition of \cite{LethamRuMcMa15}, we use a prior over rule lists to encourage sparsity. The generative process is described in Algorithm \ref{algo:rule_list}. It depends on the input prior parameters $\lambda, \eta$, and $\bm{\alpha}$, which is a user-chosen vector of size $m+1$ (as before, usually all elements in $\bm{\alpha}$ are the same and indicate pseudo-counts).
 \begin{algorithm}[t]
    \textbf{Input}:  Prior parameter $\lambda$ and $\eta$, pseudo-count $\bm{\alpha}$, observations $X = \{x_i\}$, antecedents $A= \{a_j\}$ \\
    \textbf{Output}: {Density rule list}
\begin{algorithmic}[1]
    \STATE Sample a decision list length $m \sim P(m | A, \lambda)$
    \FOR{decision list rule $j=1,\ldots, m$}
    \STATE Sample the cardinality  of antecedent $a_j$ in $d$ as $c_j \sim P(c_j| c_{<j}, A, \eta)$.
    \STATE Sample $a_j$ of cardinality $c_j$ from $P(a_j|a_{<j},c_j,A)$.
    \ENDFOR
    \FOR{observation $i=1, \ldots, n$}
    \STATE Find the antecedent $a_j$ in $d$ that is the first that applies to $x_i$.
    \STATE If no antecedents in $d$ applies, set $j=0$.
    \ENDFOR
    \caption{Density rule lists generation procedure}\label{algo:rule_list}
    \STATE Sample parameter ${\bm{\theta}} \sim$ Dirichlet ($\bm{\alpha}$) for the probability to be in each of the leaves
    \FOR{each leaf $l$ in the rule list}
    \STATE Compute volume $V_l$ according to \eqref{eq:volume}
    \STATE Compute density $f_l=\frac{\theta_l}{\mathbb{V}_l}$
    \ENDFOR
\end{algorithmic}
\end{algorithm}

For instance, in Step 1 of Algorithm \ref{algo:rule_list}'s generation process, we might find out that the rule list is of size $m=4$. Then in the Steps 2-5, we sample the cardinality of each rule, so we may find that the 4 rules are of cardinality $2, 1, 2, 3$. Then in Steps 6-9, we sample to determine which rules are actually used, for instance, the first rule might be $a_1:\{x_{14}=\textrm{blue} \;\;\&\;\; x_{21}=\textrm{large}\}$, the second rule could be $a_2:\{x_{52}=\textrm{feathered}\}$, the third rule $a_3:\{x_{13}=\textrm{10110}\;\; \&\;\; x_3=\textrm{angled}\}$, and the fourth rule $a_4:\{x_{55}={000} \;\;\&\;\; x_8=\textrm{hypothetical}\;\; \&\;\; x_{21}=\textrm{small}\}$. The rest of the feature space (that does not fall into any of these rules) would go to the default rule, again having constant density within the rule.

\noindent\textbf{Prior:} The distribution of $m$ in Step 1 is the Poisson distribution, truncated at the total number of preselected antecedents:
$$P(m|A,\lambda)=\frac{\lambda^m/m!}{\sum_{j=0}^{|A|} (\lambda^j/j!)}, m=0, \ldots, |A|.$$
When $|A|$ is huge, we can use the approximation $P(m|A, \lambda) \approx  \lambda^m/m!$, as the denominator would be close to 1.

For Step 2, we let $R_j(c_1,\ldots, c_j,A)$ be the set of antecedent cardinalities that are available after drawing antecedent $j$, and we let $P(c_j|c_{<j},A, \eta)$ be a Poisson truncated to remove values for which no rules are available with that cardinality:
$$P(c_j|c_{<j},A,\eta)=\frac{(\eta^{c_j}/c_j!)}{\sum_{k \in R_{j-1}(c_{<j},A)}(\eta^k/k!)}, \hspace{5pt} c_j \in R_{j-1}(c_{<j},A).$$
We use a uniform distribution over antecedents in $A$ of size $c_j$ excluding those in $a_{j}$,
$$P(a_j|a_{<j},c_j,A) \propto 1, \hspace{20pt} a_j \in \left\{ a \in A \setminus a_{<k}: |a|=c_j\right\}.$$
The sparse prior for the antecedent lists is thus:
$$P(d|A, \lambda, \eta)=P(m|A,\lambda) \cdot \prod_{j=1}^m P(c_j|c_{<j},A, \eta) \cdot P(a_j|a_{<j},c_j,A).$$
The prior distribution over the leaves $\bm{\theta}=[\theta_1,\ldots,\theta_m,\theta_0]$ is drawn from Dir($\bm{\alpha}_{m+1}$).
$$P(\bm{\theta}|\alpha)=\frac{1}{\Beta_{m+1}([\alpha, \cdots,\alpha])}\prod_{l=0}^m \theta_l^{\alpha-1}$$
It is straightforward to sample an ordered antecedent list $d$ from the prior by following the generative process that we just specified, generating rules from the top down.\medskip

\noindent \textbf{Likelihood:} Similar to Method I, the probability to land at any specific value within leaf $l$ is $\frac{\theta_l}{\mathbb{V}_l}$. Hence, the likelihood for the full data set is thus 
    $P(X|\bm{\theta},d)=\prod_{l=0}^{m}\left( \frac{\theta_l}{\mathbb{V}_l} \right)^{n_l}.$

\noindent \textbf{Posterior:} The posterior can be written as
\begin{eqnarray*}
P(d|A,\lambda, \eta, \alpha, X)
&\propto &\int_{\bm{\theta} \in \textrm{simplex}} P(d|A,\lambda, \eta) \cdot P(\bm{\theta} | \alpha) \cdot P(X|\bm{\theta}, d) d\bm{\theta} \\
& = & P(d|A, \lambda, \eta)   \int_{\bm{\theta} \in \textrm{ simplex}} \frac{1}{\Beta_{m+1}([\alpha, \cdots,\alpha])} 
\prod_{l=0}^m \theta_l^{\alpha-1}\left(\frac{\theta_l}{\mathbb{V}_l}\right)^{n_l} d \bm{\theta}\\
&=& P(d|A, \lambda, \eta)\frac{\prod_{l=0}^m\Gamma{(n_l+\alpha)} \mathbb{V}_l^{-n_l}}{\Gamma(\sum_{l=0}^m (n_l+\alpha))}.
\end{eqnarray*}
where the last equality uses the standard Dirichlet-multinomial distribution derivation.

\section{Numerical Methods to Optimize the Objective Function}\label{sec:optimization}

In the previous section, we have presented the posterior functions for three generative modeling methods. Since the search space of our problems is large, we use simulated annealing, a metaheuristic optimization algorithm, which allows us to approximate global solutions.
More specifically, in this section we describe a simulated annealing scheme that we implemented in order to find the optimal tree that maximizes the posterior for Method 1 and Method 2 as well as discuss Method 3's optimization details.

\textbf{Simulated annealing for tree-based methods:}
A successful simulated annealing scheme requires us to create a useful definition of a neighborhood. We define our neighborhood such that each move explores a neighboring tree  where we are able to extend or shrink the tree.

At each iteration we need to determine which neighboring tree to move to.
To decide which neighbor to move to, we fix a parameter $\epsilon>0$ beforehand, where $\epsilon$ is small, approximately 0.01 in our experiments. $\epsilon$ is the probability that we will perform a structural change to jump out of a possible local minimum. All other actions below are taken with equal probability. 
Thus, at each time, we generate a number from the uniform distribution on $(0,1)$, then either:\\
1. (Shrink at leaf) If the number is smaller than  $\frac{1-\epsilon}{4}$, we select uniformly at random a ``parent'' node which has leaves as its children, and remove its children. This is always possible unless the tree is the root node itself, in which case we cannot remove it and this step is skipped.
\\ 
2. (Expand) If the random number is between $\frac{1-\epsilon}{4}$ and $\frac{1-\epsilon}{2}$, we pick a leaf randomly and a feature randomly. If it is possible to split on that feature, then we create children for that leaf. (If the feature has been used up by the leaf's ancestors, we cannot split, and we then skip this round.)
\\ 
3. (Regroup) If the random number is between $\frac{1-\epsilon}{2}$ and $\frac{3(1-\epsilon)}{4}$, we pick a node randomly, delete its descendants, and split the node, creating two child nodes where the splitting is done on subsets of the node's feature values. Sometimes this is not possible, for example if we pick a node where all the features have been used up by the node's ancestors, or if the node has only one category. In that case we skip this step. \\
4. (Merge sibling nodes) If the random number is between $\frac{3(1-\epsilon)}{4}$ and $(1-\epsilon)$, we choose two nodes that share a common parent, delete all their descendants and merge the two nodes (e.g., black, white, red, green becomes black-or-white, red, green).\\
5. (Structural change) 
If the random number is more than $1-\epsilon$, we perform a structural change operation where we remove all the children of a randomly chosen node of the tree.

Please see Algorithm \ref{algo:treeSA} for the pseudo-code of the simulated annealing procedure. The last three actions avoid problems with local minima. The algorithms can be warm-started using solutions from other algorithms, e.g., DET trees. We found it useful to occasionally reset to the best tree encountered so far or the trivial root node tree. 

 \begin{algorithm}[t]
    \textbf{Input}:  Prior parameters, $\theta$, maximum number of iterations, $N$, $\epsilon$, cooling schedule Cool(iteration) for the simulated annealing\\
    \textbf{Output}: {Optimal density tree}
\begin{algorithmic}[1]
    \STATE Initialize the initial tree $T$ to be a single node, compute the posterior using the objective function and the prior parameters. Set iteration number to be $0$.
    \WHILE {iteration number $< N$}
    \STATE Draw a random number $r$, from $Uni(0,1)$.
    \IF {$r<\frac{1-\epsilon}{4}$}
    \STATE Perform shrink at leaf operation.
    \ELSIF {$r<\frac{2(1-\epsilon)}{4}$}
    \STATE Perform expand operation
    \ELSIF {$r <\frac{3(1-\epsilon)}4$}
    \STATE Perform regroup operation.
    \ELSIF {$r<1-\epsilon$}
    \STATE Perform merge sibling nodes operation.
    \ELSE 
    \STATE Perform structural change.
    \ENDIF
    \STATE Compute the objective value for modified tree.
    \IF {a better objective value is obtained than current best}
    \STATE Update $T$ to be the current tree.
    \ENDIF
    \IF {the current tree is worse than the current best tree $T$}
    \STATE With probability defined by the cooling schedule Cool(iteration), update $T$ to be the current tree. This will always be a small probability.
    \ENDIF
   \ENDWHILE
   \STATE Return $T$
    \caption{Simulated annealing for tree-based methods}\label{algo:treeSA}
\end{algorithmic}
\end{algorithm}

\textbf{Sparse Density Rule List Optimization:}
To search for optimal density rule lists that fit the data, we use local moves (adding rules, removing rules, and swapping rules) and use the Gelman-Rubin convergence diagnostic applied to the log posterior function.

A technical challenge that we need to address in our problem is the computation of the volume of a leaf. Volume computation is not needed in the construction of a decision list classifier like that of \cite{LethamRuMcMa15} but it is needed in the computation of density list. There are multiple ways to compute the volume of a leaf of a density rule list.\\ 
\textit{Approach 1: Analytical Computation.} Use the inclusion-exclusion principle to directly compute the volume of each leaf. Consider computing the volume of the $i$-th leaf in a density rule list.  Let $V_{a_i}$ denote the volume induced by the rule $a_i$, that is the number of points in the domain that satisfy $a_i$. To belong to that leaf, a data point has to satisfy $a_i$ and not earlier rules $a_{<i}$. Hence the volume of the $i$-th leaf is equal to the volume obeying $a_i$ alone, minus the volume that has been used by earlier rules. Using notation $a_k^c$ to denote the complement of condition $a_k$, we have the following:
\begin{align}\nonumber
\mathbb{V}_i&=V_{a_i \wedge \bigwedge_{k=1}^{i-1} a_k^c} \;\;\textrm{ ($i$ is in $a_i$ and in the complement of all previous rules)}\\\nonumber
&=V_{a_i}-V_{a_i \wedge (\bigvee_{k=1}^{i-1} a_k)}
\\\nonumber
&=V_{a_i}-V_{ (\bigvee_{k=1}^{i-1}a_i \wedge a_k)}
\\&=V_{a_i}-\sum_{k=1}^{i-1}(-1)^{k+1}\sum_{1\leq j_1 \leq \ldots j_k \leq n} V_{a_i\wedge a_{j_1} \ldots \wedge a_{j_k}},\label{eq:volumedecompose}
\end{align}
where the last expression is due to the inclusion-exclusion principle and it only involves the volume resulting from conjunctions.
The volume resulting from conjunctions can be easily computed from data. Without loss of generality, suppose we want to compute the volume of $V_{a_1 \wedge \ldots \wedge a_k}$. For each feature that appears, we examine if there is any contradiction between the rules; for example, if feature 1 is present in both $a_1$ and $a_2$, where rule $a_1$ specifies feature 1 to be 0 whereas $a_2$ specifies feature 1 to be 1, then we have found a contradiction and the volume of the intersection of $a_1$ and $a_2$ should be 0. If there is no contradiction, then the volume is equal to the product of the number of distinct categories of all the features that are not used. If all features are used, then the volume is 1. By using the inclusion-exclusion principle, we reduce the problem to just computing a volume of conjunctions as in \eqref{eq:volumedecompose}. Note that for this approach, we still need to iterate over all conjunctions for each volume computation, which can be computationally expensive.
\textit{Approach 2: Uniform Sampling.} Create uniform data over the whole domain, and count the number of points that satisfy the antecedents. This approach would be expensive when the domain is huge but easy to implement for smaller problems. \\
\textit{Approach 3: MCMC.} Use an MCMC sampling approach to sample the whole domain space. This approach is again not practical when the domain size is huge as the number of samples required will increase exponentially due to curse of dimensionality.

We use the analytical computation approach 1 in our implementation.\\

 
 Some works \citep{AngelinoEtAl2018,ErtekinRu18}
 have achieved provable optimality on minimization of objectives over a set of pre-mined rules, though for supervised classification (not density estimation). They have also noted that randomized methods, such as  that of \cite{YangRuSe16} (or
 those considered in the present work), tend to produce models that are close to these optimal solutions, leading us to believe that our search methods may actually achieve close-to-optimal solutions fairly often, particularly for problems where there may be a large ``Rashomon'' set of good models \citep{SemenovaRuPa2022,FisherRuDo19}. None of these earlier works consider density estimation, but nonetheless, we have reason to hypothesize that simulated would also yield near-optimal models for density estimation.


\section{Experiments}\label{sec:experiments}

Our experimental setup is as follows. We considered five methods: the leaf-sparse density tree, the branch-sparse density tree, the sparse density rule list, and regular histograms and density estimation trees (DET) \citep{wu2018density}. 
Note that the implementation of DET is meant for continuous variables, but we use it anyway for comparison.
To our knowledge, these methods can serve to represent the full set of logic-based, high dimensional density estimation methods. To ascertain uncertainty, we split the data in $5$ folds and assessed test log-likelihood (i.e., out-of-sample performance) and sparsity of the trees for each method on every fold. A model with fewer bins and higher test likelihood is a better model. 
  
Let us discuss how the baselines were implemented.  For the standard high-dimensional histogram baseline, we treated each possible set of feature values (e.g., $x_{.1}=1, x_{.2}=0, ..., x_{.10}=1$) as a separate bin. (We call a set of feature values a \textit{configuration}; it is a point in our feature space.) Histograms have the disadvantage that they create a large number of bins and thus may not generalize well to the test set; they are also not interpretable, since they cannot be visualized in a tree or list.

DET was designed for continuous data, which meant that the computation of volume needed to be adapted for discrete data -- it is the number of configurations in the bin, rather than the lengths of each bin multiplied together. Thus, we used our own computations for the density in each leaf following volume computations in \eqref{eq:volume}.

The DET method has two parameters, the minimum allowable support in a leaf, and the maximum allowable size of a leaf. We originally planned to use a minimum of 0 and a maximum of the size of the full data set, but the algorithm often produced trivial models when we did this (i.e., models with one leaf). Therefore we tried values  $\left\{ 1,3,5\right\}$ for the minimum size of a leaf and $\left\{ 5, 10, 50, 100, \lfloor\frac{n}{10}\rfloor, \lfloor\frac{n}{5}\rfloor \right\}$ for the maximum size of a leaf, where $n$ is the number of training data points. We use nested cross-validation over 5 folds. For each fold, we optimize parameters for validation log-likelihood and report out-of-distribution log-likelihood. DET has the disadvantage of being a greedy algorithm and the available implementation of DET is not designed for categorical data (see Appendix \ref{appendix:det_implementation}), thus DET may not produce trees that are as useful or sparse as those from Methods I, II, or III.

For the data sets in the following two sections for the leaf-sparse density tree model (Method I), the mean of the Poisson prior was chosen from the set $\left\{ 5,8, 10\right\}$ using nested cross validation. For the branch-sparse density tree model (Method II), the parameter to control the number of branches was chosen from the set  $\left\{ 2,3\right\}$. 
$\alpha$ was set to be 2 for the experiment. This corresponds to a 
pseudocount of 2 data points in each bin (to prevent bins with 0 data points). For the sparse density rule list (Method III), the parameters $\lambda,\eta$ and $\alpha$ were chosen among $[3,5,7], 1$, and $1$ respectively. We provide a summary of parameters and their suggested values  in Appendix \ref{appendix:user_guide}.

\begin{figure*}[t]
\subfloat[]{\includegraphics[width = 2in]{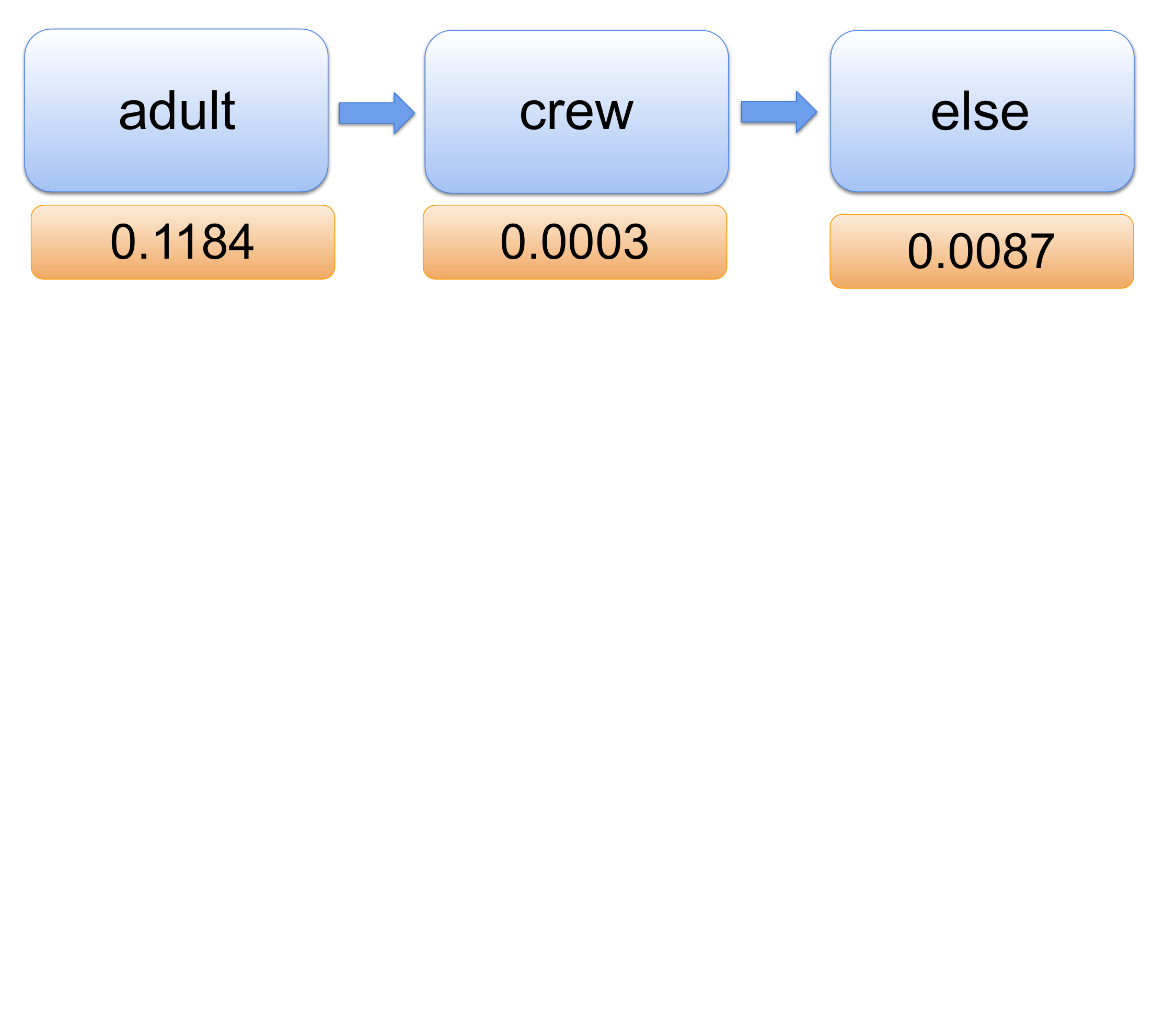}} \hfill
\subfloat[]{\includegraphics[width = 2in]{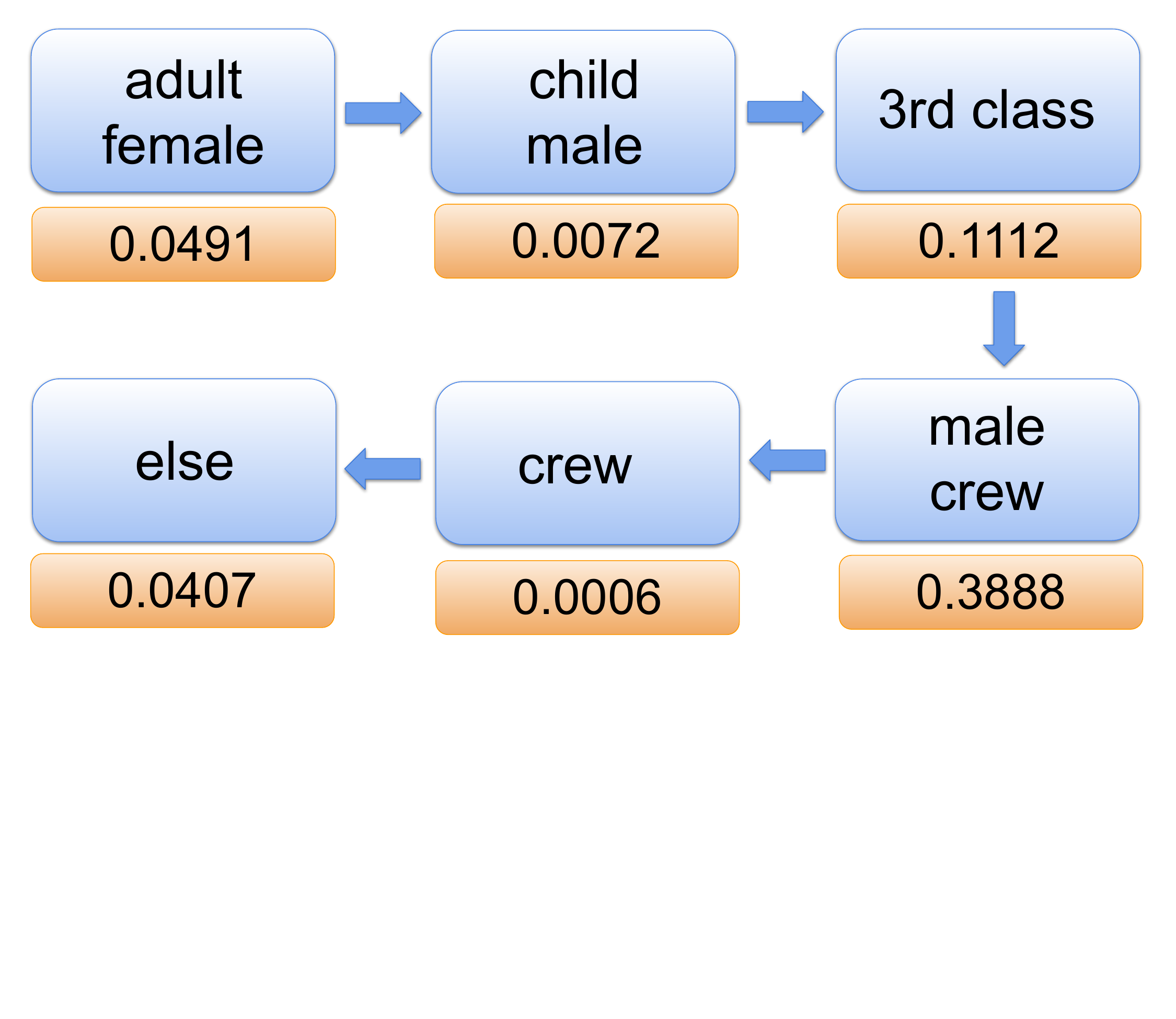}} \hfill
\subfloat[]{\includegraphics[width = 2in]{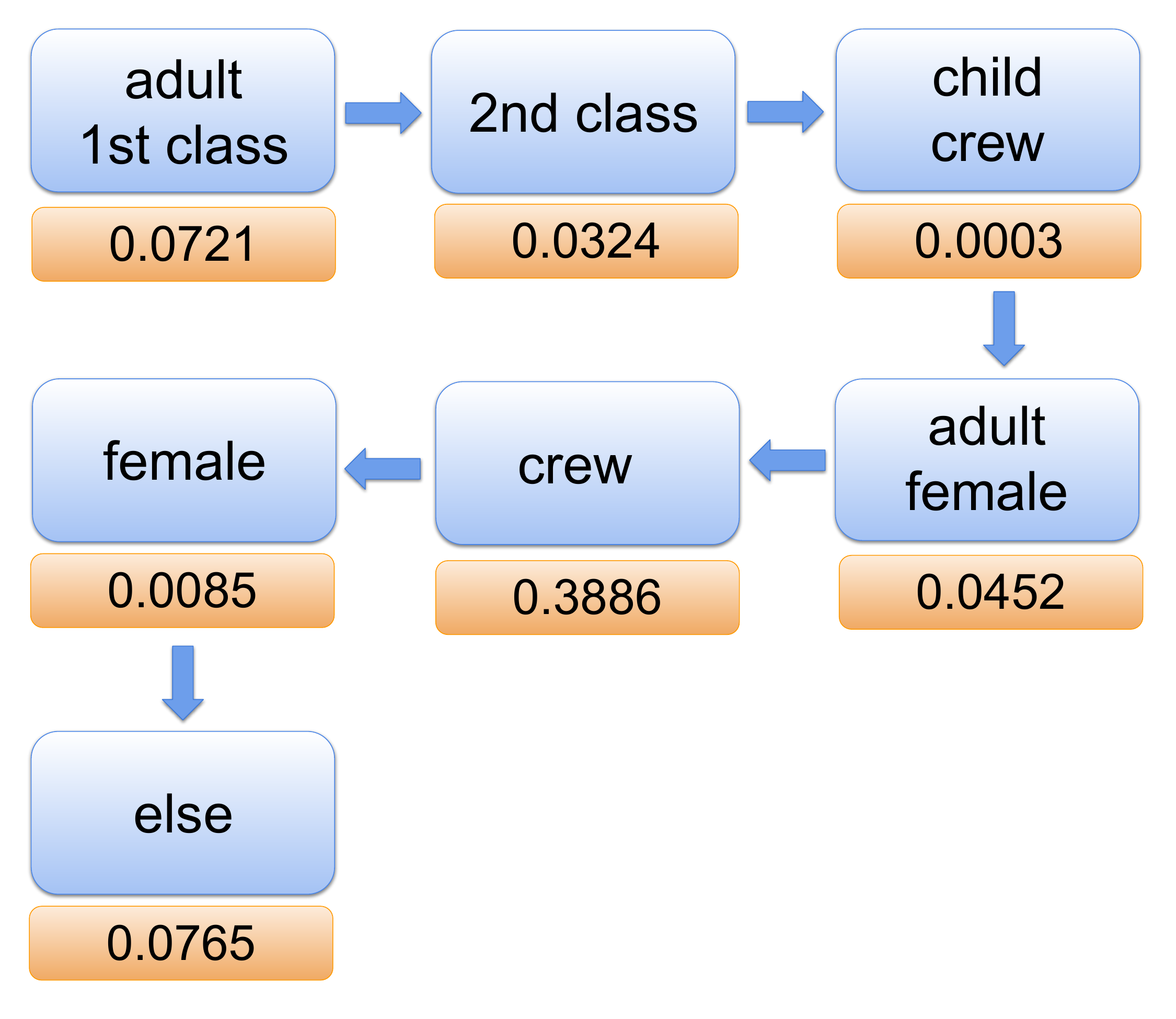}}
\caption{Density rule lists with different parameter $\lambda$ that indicates preferred list length. $\lambda$ is set to 2 for (a), 4 for (b), and 7 for (c). Parameters $\eta$ and $\alpha$ were chosen as 2 and 1 respectively. These density lists were chosen based on the maximum log-likelihood over 5 repeats. Each arrow represents an ``else if" statement. E.g., for (b) if the passenger is adult and female, then density is constant with respect to other variables at 0.0491, else if passenger is a child and male, density is constant at 0.0072, etc.}
\label{fig:titanic_length}
\end{figure*}

\subsection{Titanic Data Set}

As discussed earlier, a sparse density tree or list would help us understand the distribution of people on board the Titanic.
The Titanic data set has an observation for each of the 2201 people aboard the Titanic. There are 3 features: gender, whether someone is an adult, and the class of the passenger (first class, second class, third class, or crew member).

\begin{figure*}[t]
\centering
\subfloat[]{\includegraphics[height = 1.8in]{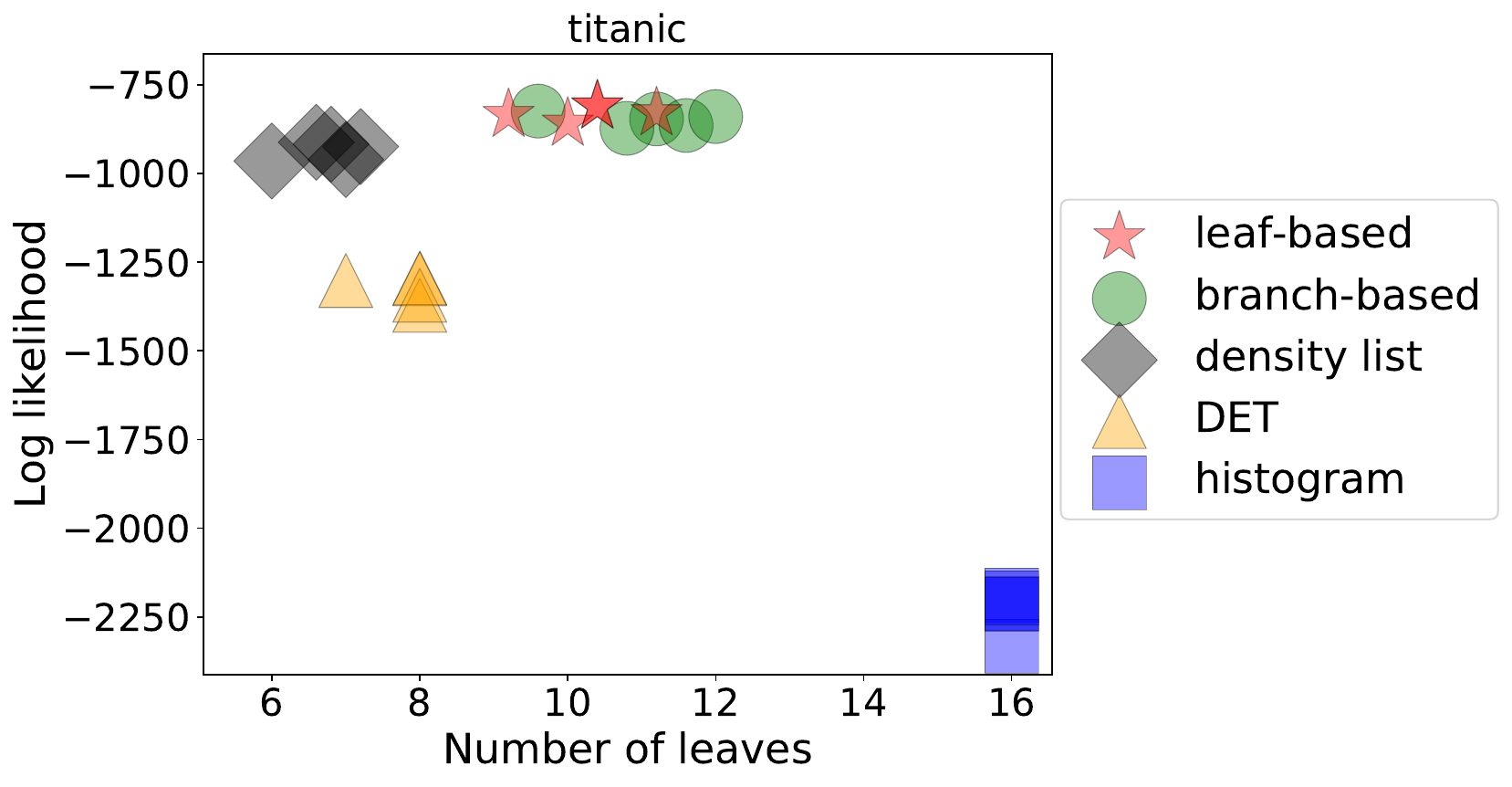}}  
\subfloat[]{\includegraphics[height = 1.8in]{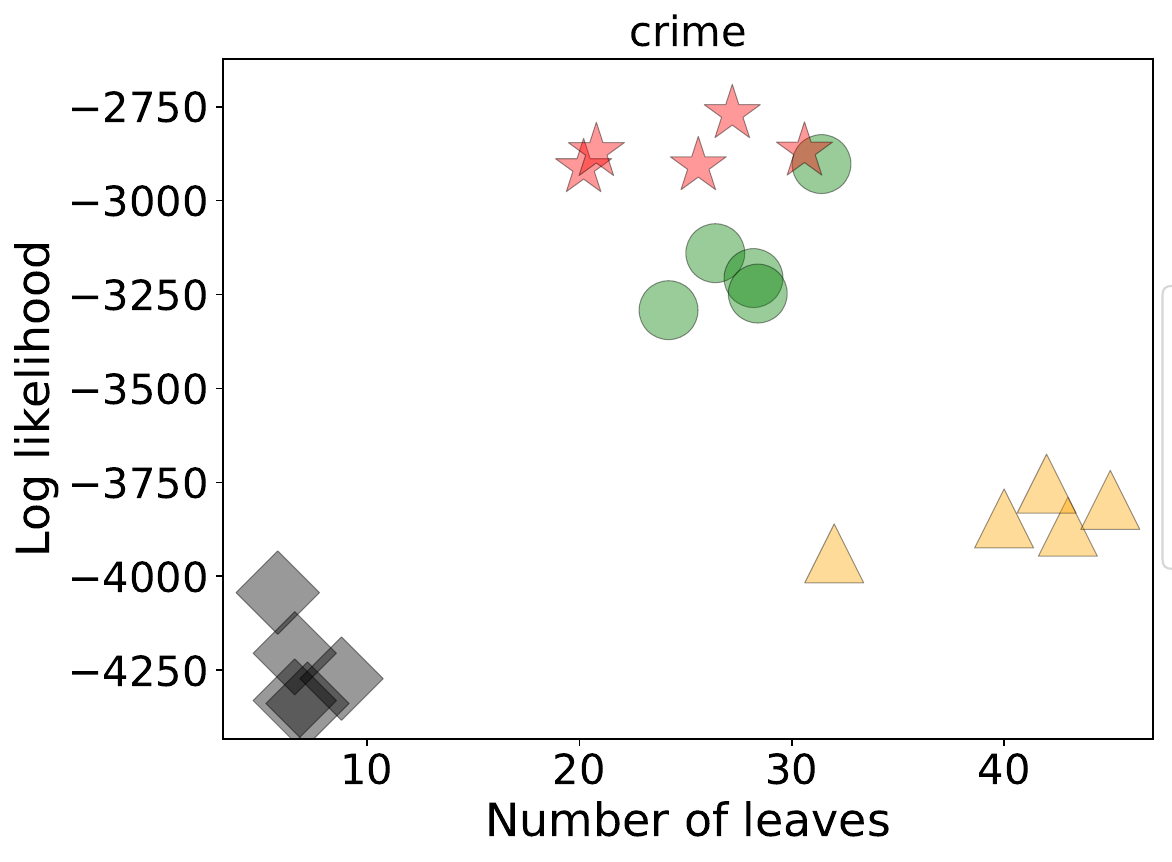}} 
\caption{Performance comparison between our methods and baselines for Titanic (a), Crime (b) data sets.}
\label{figure345}
\end{figure*}

Figure \ref{figure345} shows the results, both for out-of-sample log-likelihood (on y-axis) and sparsity (on x-axis), for each method, for each of the 5 test folds. The histogram method had the most leaves (by design), and thus tended to overfit. Our methods performed well, arguably the sparse density rule list method performed slightly better in the likelihood-sparsity tradeoff. 
In general, the results are consistent across folds: the histogram produces too many bins, the sparse density rule list method and density tree methods performs well, and DET has worse log-likelihood. 

Figure \ref{fig:example_titanic} shows one of the density models generated by the leaf-sparse density tree method. The reason for the split is clear: there were fewer children than adults, the distributions of the males and females were different (mainly due to the fact that the crew was mostly male), and the volume of crew members was very different than the volume of first, second, and third class passengers. 

 Figure \ref{fig:titanic_length} shows density rule lists for the Titanic data set, by choosing parameter $\lambda$, which indicates preferred list length, from set $\{2,4,7\}$, we change the density rule lists as well as its length. Lower values of the parameter lead to shorter density rule lists, while larger preferred lengths correspond to longer density rule lists.

\begin{figure}[t]
\centering
\includegraphics[width=\textwidth]{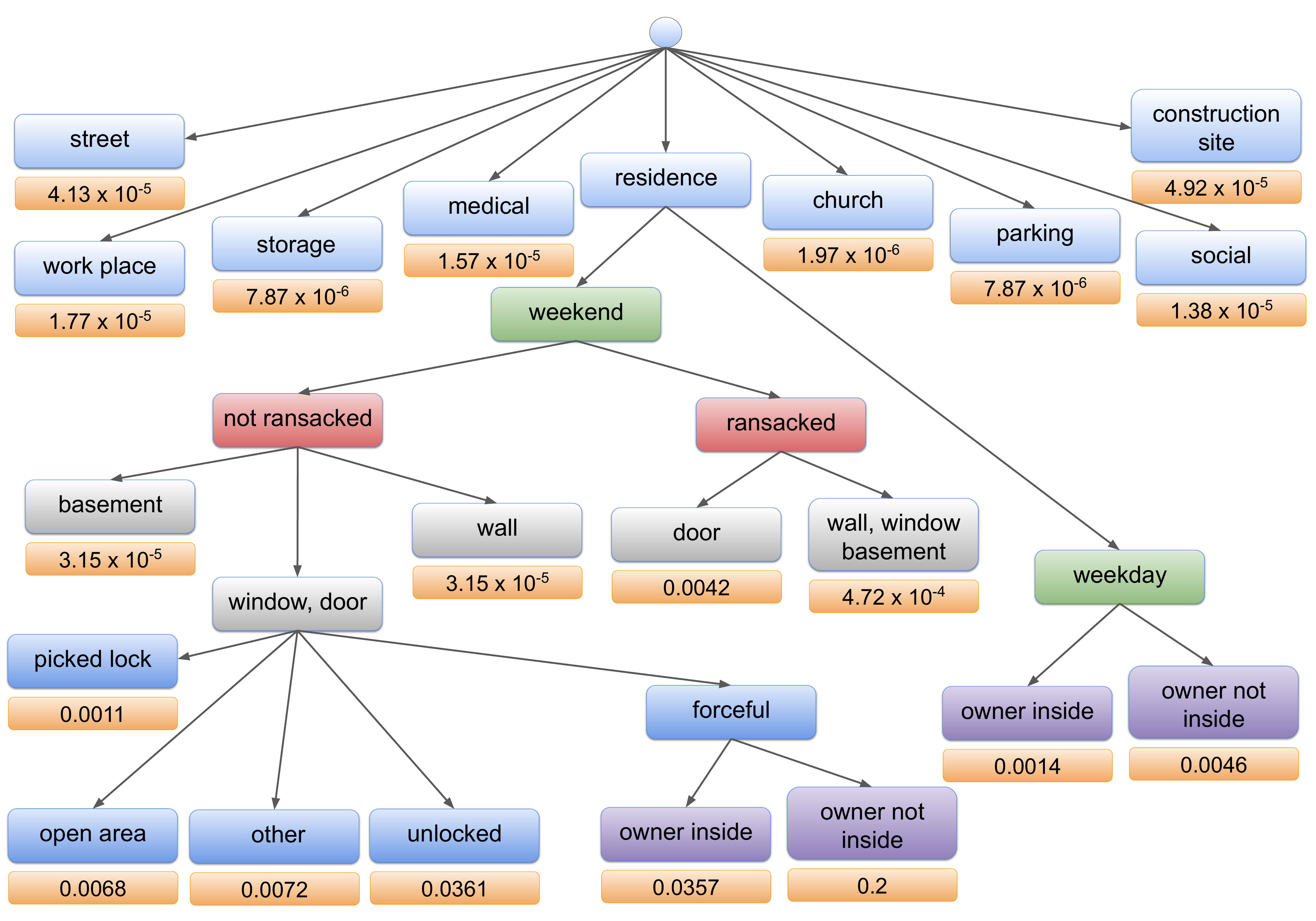}
\caption{Leaf-sparse density tree representing the crime data set. Density is constant in each leaf. Different node colors represent different features in the data set. This tree contains 20 leaves. It took around 1.4 seconds to create the tree and around 4 seconds to run the validation process. \label{fig:crimetree}}
\end{figure}

\subsection{Crime Data Set}
The housebreak data used for this experiment are from the Cambridge Police Department, Cambridge, Massachusetts. 
The motivation is to understand the common types of modus operandi (M.O.) characterizing housebreaks, which is important in crime analysis. The data consist of 3739 separate housebreaks occurring in Cambridge between 1997 and 2012 inclusive. 
The 6 categorical features for the crime data set are as follows:
1) Location of entry: ``window,'' ``door,'' ``wall,'' and ``basement.''
2) Means of entry: ``forceful'' (cut, broke, cut screen, etc.), ``open area,'' ``picked lock,'' ``unlocked,'' and ``other.''
3) Whether the resident is inside.
4) Whether the premise is judged to be ransacked by the reporting officer. 
5) Whether the entry happened on ``weekday'' or ``weekend.'' 
6) Type of premise. The first category is ``residence'' (including 
apartment, residence/unk., dormitory, single-family house, two-family house, garage (personal), porch, apartment hallway, residence unknown, apartment basement, condominium). The second category is non-medical, non-religious ``work place'' (commercial unknown, accounting firm, research, school). The third group ``medical'' consists of halfway houses, nursing homes, medical buildings, and assisted living. The fourth group ``parking'' consists of parking lots and parking garages, and the fifth group ``social'' consists of YWCAs, YMCAs, and social clubs. The last groups are ``storage,'' ``construction site,'' ``street,'' and ``church,'' respectively.

The experiments in Figure \ref{figure345}(b) show that our approaches dominate DET for the Crime data set  and are sparser than DET trees. The standard histogram's results were not reported since they involve too many bins (1440) to fit on the figure, and are thus not competitive.

\begin{figure}[t]
   \centering
   \begin{tabular}{@{}c@{\hspace{.5cm}}c@{}}
     \includegraphics[width=0.7\textwidth]{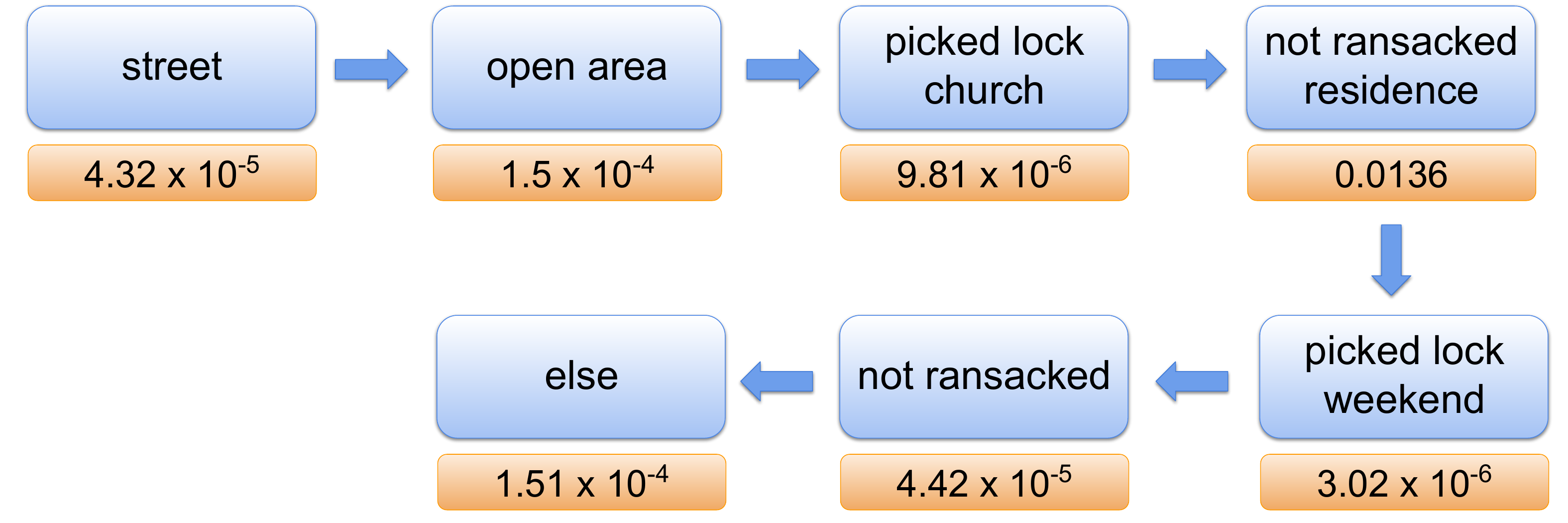}
   \end{tabular}
 \caption{List representing the crime data set. Each arrow represent an ``else if" statement.}
 \label{fig:crimelist}
\end{figure}

One of the trees obtained from the leaf-sparse density tree method 
is in Figure \ref{fig:crimetree}.
It states that most burglaries happen at residences -- the non-residential density has values less than $2 \times 10^{-4}.$
Given that a crime scene is a residence, most crimes happened on weekends. For residential crimes, burglary is more likely to happen when the owner is not inside (density 0.0046 if weekday and 0.2 if weekend, the premise is judged to be not ransacked and there is forceful entry through the door or window). When the premise is judged to be ransacked, the crime is more likely to happen with the door as the location of entry (density 0.0042) compared to wall, window, and basement (density $4.72 \times 10^{-4}).$ On weekends, for residential and not-ransacked premises, doors and windows are more common locations of entry. 
If the entry is not forceful, unlocked windows and doors are the most common means of entry (density is 0.0361). If the means of entry is picked lock, the density is 0.0011 and if the area is open, the density is 0.0068.

%
%
Important aspects of the modus operandi are within the leaves of the tree, for instance, that the owner of a residence is not inside, and the house was not ransacked and the entry was forceful through the door or window. If this approach would have been performed using a regular histogram, it would require $1440$ different markers (discrete states), whereas the crime tree groups the crimes into just $20$ bins. 

These types of results can be useful for crime analysts to assess whether a particular modus operandi is unusual. For instance, according to the tree, it is clearly more unusual for the owner to be inside during the break-in, as shown by the smaller density values in the leaves when the owner is inside. Also, according to the densities in the leaves, it is more common for the means of entry to be forceful, and for the location of entry to be windows and doors. A density list for these data is presented in Figure \ref{fig:crimelist}. The preferred length of the list was chosen from the set $\{3,5,7\}$.



\section{Empirical Performance Analysis}\label{sec:perf_analysis}
The experiments below are designed to provide insight into how the methods operate.

\begin{figure}[t]
\subfloat[]{\includegraphics[width = 2in]{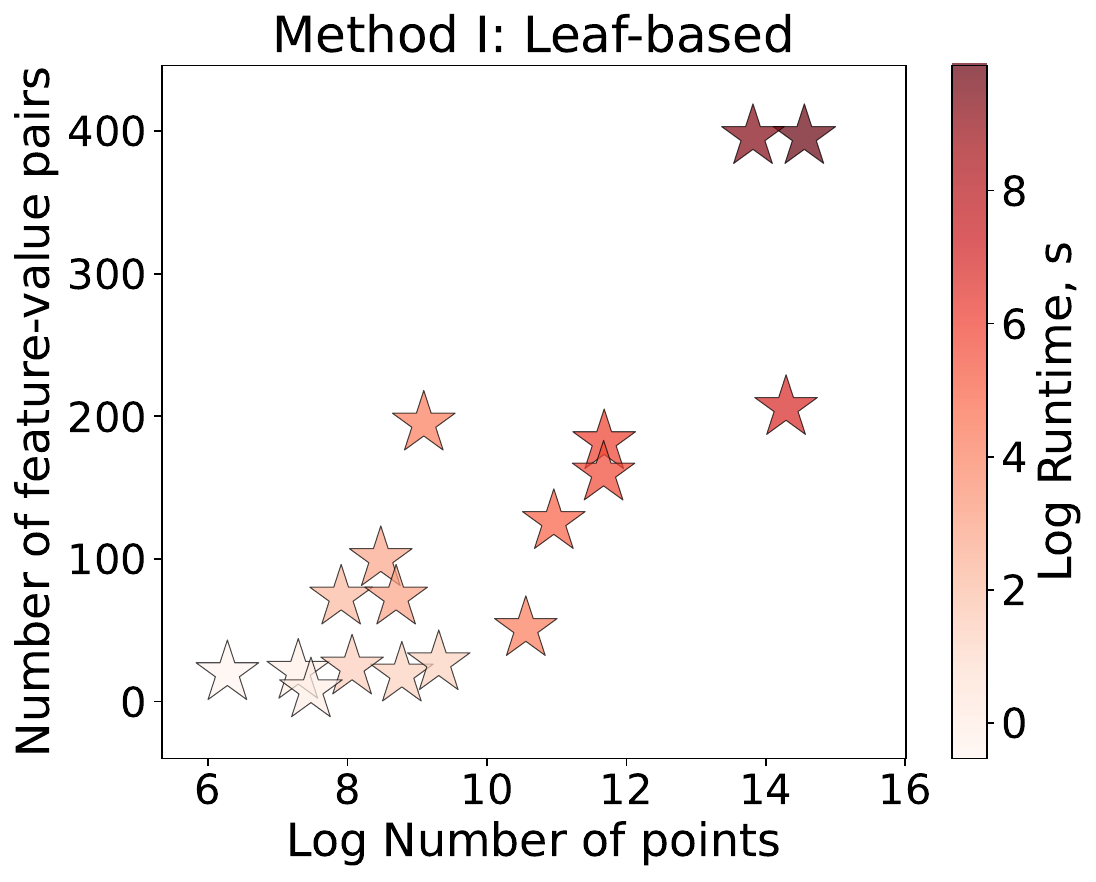}} \hfill
\subfloat[]{\includegraphics[width = 2in]{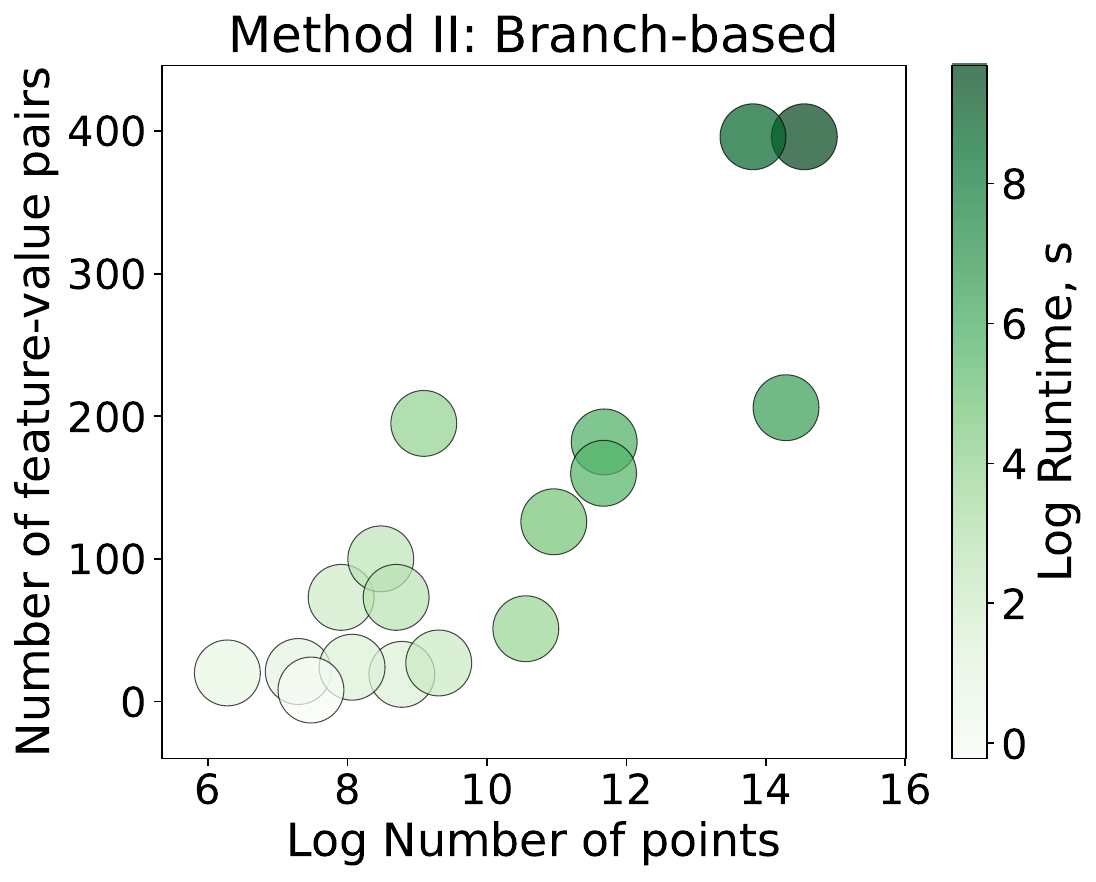}} \hfill
\subfloat[]{\includegraphics[width = 2in]{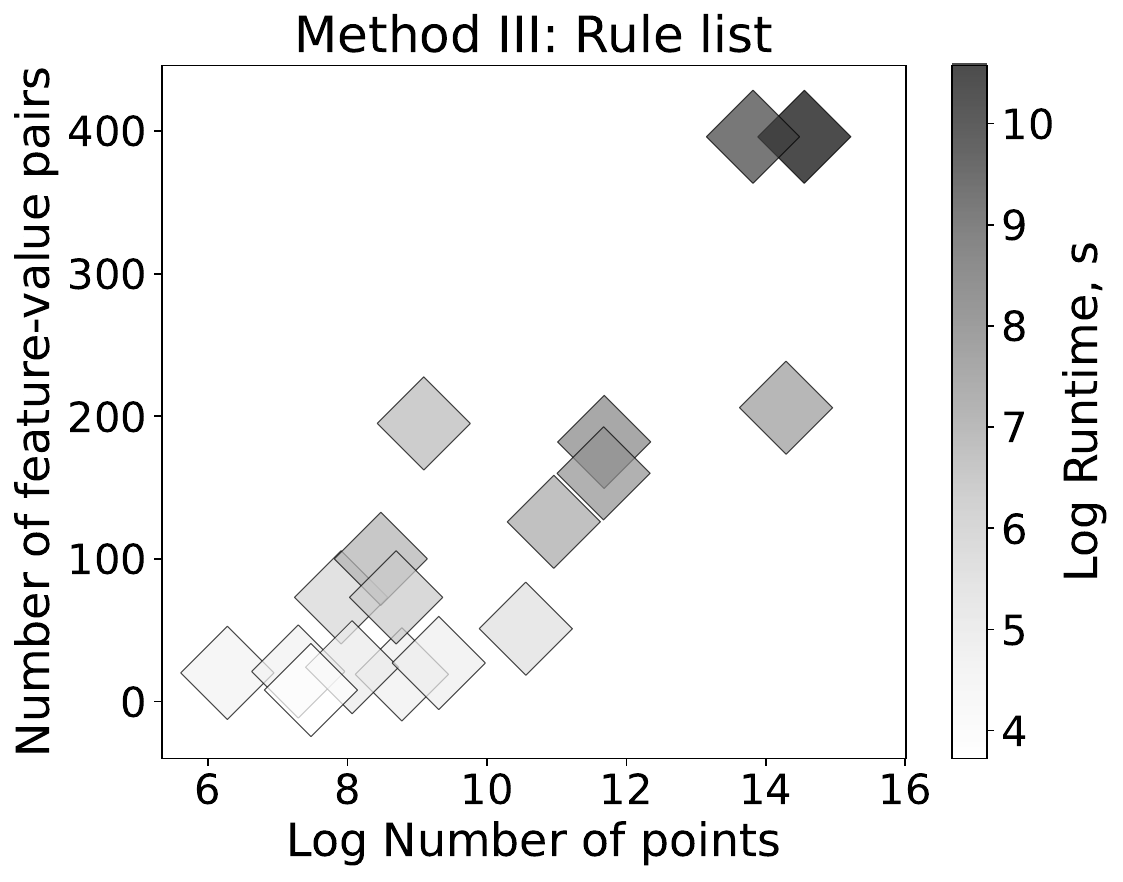}}
\caption{Algorithm run time for all data sets as a function of data set complexity (the log of the number of samples and the number of feature-value pairs) for (a) leaf-sparse density tree, (b) branch-sparse density tree, and (c) sparse density rule list estimation methods. Please refer to Table \ref{table:runtime_results} for more details.}
\label{figure:runtime}
\end{figure}

\textbf{Run Time Analysis:}
We studied the performance of the density estimation methods on data sets of different complexity. We chose 17 data sets, including financial data sets (Bank-Full, Telco Customer Churn, HELOC), recidivism risk score data (COMPAS), UCI repository data sets (e.g., Car, Mushroom, US Census data), and detection labels of COCO stuff$+$thing data. The complexity of the data set, in this case, is defined based on the number of samples and/or the number of feature-value pairs (the sum of categories for each feature). For the considered data sets, the number of samples ranged from 625 to around 2.5 million. The number of feature-value pairs ranged from 9 to around 400 and there were from 3 to 68 features. Please see Table \ref{table:datasets_description} in Supplementary Section \ref{appendix:runtime} for data set statistics and pre-processing steps taken.

For data sets with less than 100k samples and less than 200 feature-value pairs, the tree-based methods (Method I and II) performed in under a minute and the rule list method (Method III) in under 7 minutes. The most complicated data set, in terms of both the number of samples and feature-value pairs, is the US Census data set ($\sim$2.09M training samples, $\sim$400 feature-value pairs) which took around 1.75 hours for Method I, 2.5 hours for Method II, and 8$\frac{2}{3}$ hours  for Method III (note that majority of the time for Method III went into data loading and pre-processing). In our implementation, we represent data through bit vectors, thus an increase in the number of samples causes a relatively smaller increase in the run time compared to the run time increase caused by a larger number of feature categories. We also found that the leaf-sparse density tree method (Method I) performed fastest on average for all data sets considered. In Figure \ref{figure:runtime} we provide a visualization of the time taken to estimate the density of each data set given its complexity for all three methods. Table \ref{table:runtime_results} in Supplementary Section \ref{appendix:runtime} shows more details on the timing. All time measurements are averaged over 5 repeats.

\textbf{Recovering a Sparse Tree that Generates Data:} This experiment is a test to see whether we can recover a tree that actually generates a data set.
Specifically, we generated a data set that arises from a tree with 6 leaves, involving 3 features. The data consists of 1000 data points, where 100 points are tied at value (1,2,1), 100 points are at (1,2,2), 100 points are at (2,1,1), 400 points are at (2,1,2), and 300 points are at (2,2,2). 

\begin{figure}[t]
\centering
\subfloat[]{\includegraphics[width = 6cm]{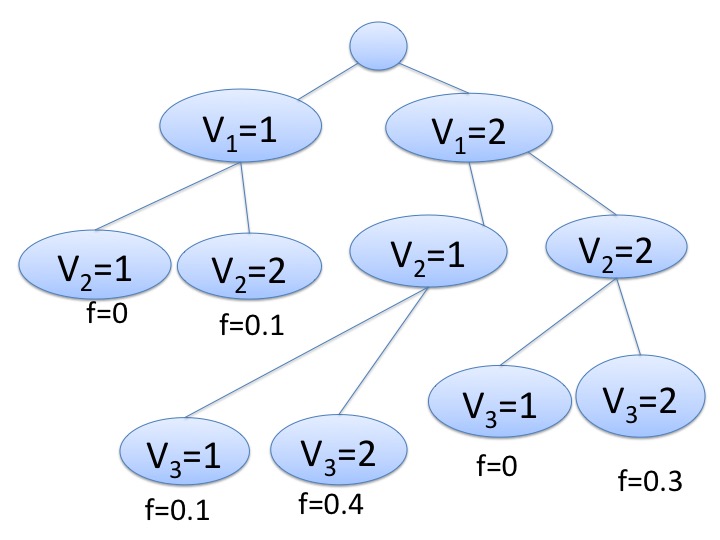}} \hspace{1cm}
\subfloat[]{\includegraphics[width = 6cm]{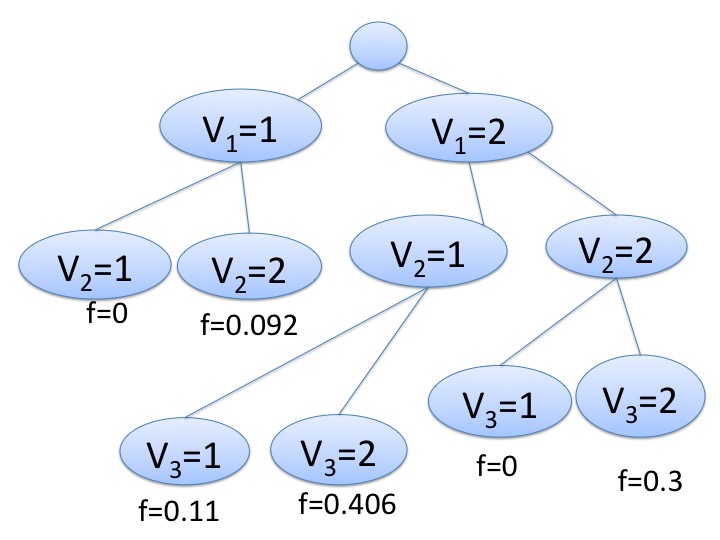}}\\
\subfloat[]{\includegraphics[height = 1.8in]{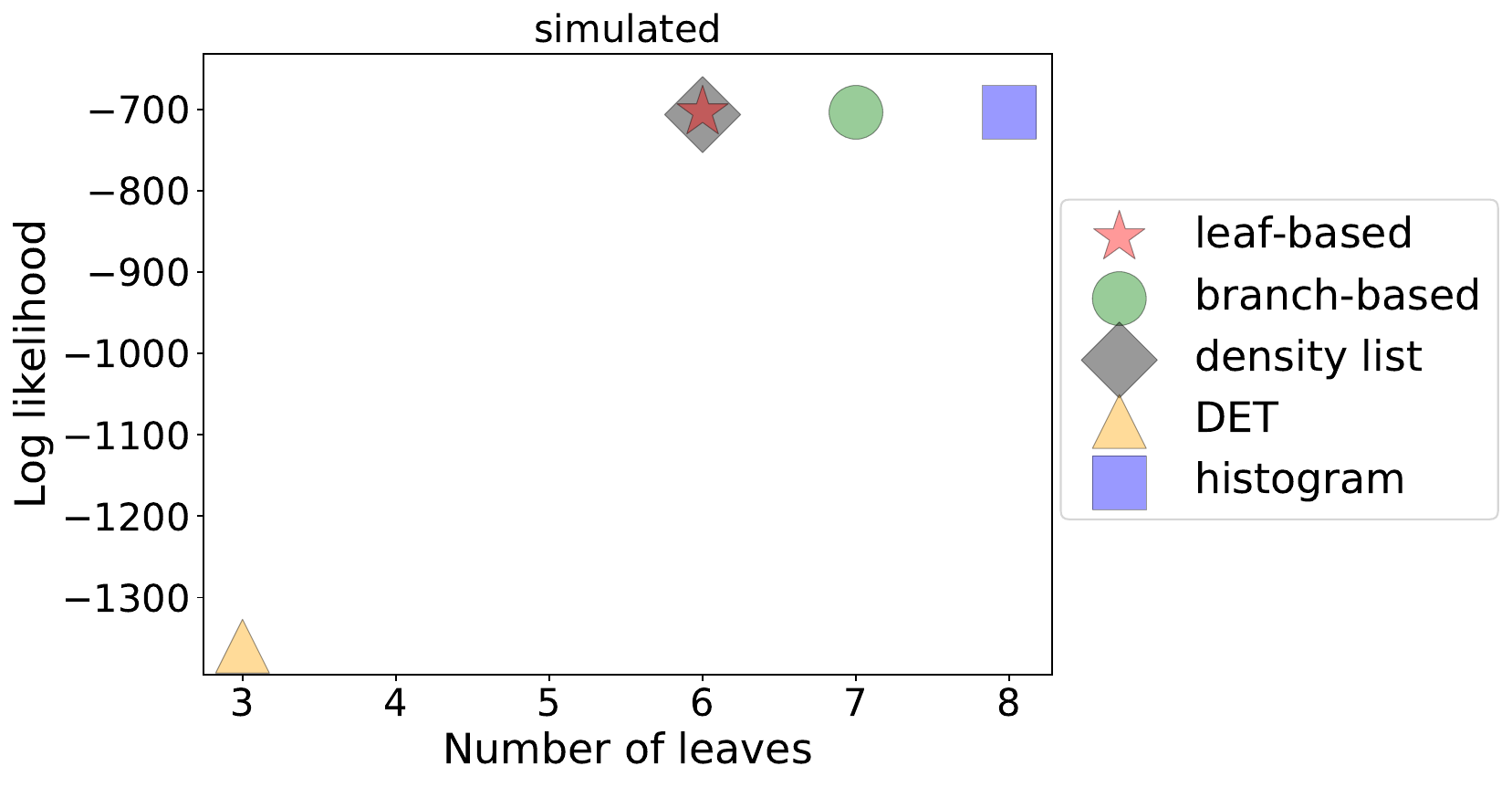}}
\caption{(a) Tree diagram for sparse tree data set. (b) Leaf-sparse density tree model that recovers the true generative structure on half of the data set. (c) Performance vs sparsity on the sparse tree data set.}\label{Fig:simdata}
\end{figure}

We trained the methods on half of the data set and tested on the other half. Figure \ref{Fig:simdata}(c) shows the scatter plot of out-of-sample performance and sparsity. This is a case where the DET failed badly to recover the true model. It produced a model that was too sparse, with only $3$ leaves. The leaf-sparse density tree method recovered the full tree.  Figure \ref{Fig:simdata}(a) is the density tree that generated the data set and we present the tree that we obtained in Figure \ref{Fig:simdata}(b) where it is recovered automatically. 



\section{Consistency}\label{sec:consistensy}
A consistent method has estimates that converge to the real densities as the size of the training set grows. Consistency of conventional histograms is well studied \citep[for example][]{abou1976conditions,devroye1983distribution}. More generally, consistency for general rectangular partitions has been studied by   \citet{zhao1991almost} and   \citet{lugosi1996consistency}.

Typical consistency proofs \citep[e.g.,][]{citeulike:3463149,Ram:2011:DET:2020408.2020507} require the leaf diameters to become asymptotically smaller as the size of the data grows. In our case, if the ground truth density is a tree, we do \textit{not} want our methods to asymptotically produce smaller and smaller bin sizes, we would rather they reproduce the ground truth tree. This means we require a new type of consistency proof. 





\noindent\textbf{Definition 1}: Trees have a single root and there are conditions on each branch. A density value, $f_l$ is associated with each leaf $l$ of the tree. 

\noindent\textbf{Definition 2}: Two trees, $T_1$ and $T_2$ are \textrm{equivalent} with respect to density $f$ if they assign the same density values to every data point on the domain, that is $f_{T_1}(x)=f_{T_2}(x)$, for all $x$. We denote the class of trees that are equivalent to $T$ as $[T]_f$.

\noindent\textbf{Theorem 1}: 
Let $\Theta$ be the set of all trees.
Consider these conditions:
\begin{enumerate}
\item $T_n \in \argmax_T Obj(T)$. The objective function can be decomposed into $\textrm{Obj}(T)=\ln q_n(T|X)+\ln g_n(T|X)$ where $\argmax_T \left[ \ln q_n(T|X)+ \ln g_n(T|X) \right]\equiv \argmax_T \ln g_n(T|X)$ as $n \rightarrow \infty$.
\item $\ln g_n(T|X)$ converges  in probability, for any tree $T$, to the empirical log-likelihood that is obtained by the maximum likelihood principle, $\hat{l}_n(T|X)=\frac{1}{n}\sum_{i=1}^n \ln \hat{f}_n(x_i|T)$.
\item $\sup_{T \in \Theta}|\hat{l}_n(T|X)-l(T)| \xrightarrow{P} 0$ where  $l(T)=\mathbb{E}_x(\ln(f(x|T)))$. 
\item $T^*_{\textrm{MLE}}\in \argmax_Tl(T)$ is unique up to equivalence among elements of $[T^*_{\textrm{MLE}}]_f$.
\end{enumerate}
If these conditions hold, then the trees $T_n$ that we learned, $T_n\in \argmax_T \textrm{Obj}(T)$, obey $T_n\in [T^*_{\textrm{MLE}}]_f$ for $n >M$ for some $M$.

The proof of the result is as follows:
The first condition and the second condition are true any time we use a Bayesian method. They are also true any time we use regularized empirical likelihood where the regularization term's effect fades with the number of observations.
Note that the third condition is automatically true by the law of large numbers. The last condition is not automatically true, and requires regularity conditions for identifiability.
The result states that our learned trees are equivalent to maximum likelihood trees when there are enough data. 

From definition of $T_n$, $T_n$ is an optimal value of the log-objective function and hence it is also an optimal solution to $g_n$ as $n$ is sufficiently large due to the first condition. We have that 
\[ \ln \textrm{Obj}(T_n|X)-\ln \textrm{Obj}(T^*_{\textrm{MLE}}|X) \geq 0 \]
by definition of $T_n$ as the maximizer of \textrm{Obj}. Because \textrm{Obj} becomes close to $g_n$, we have that
\begin{equation}\label{thmeqn1}
\ln g_n(T_n|X) - \ln g_n(T^*_{\textrm{MLE}}) \geq 0
\end{equation} 
as $n$ is sufficiently large.

From Condition 2, we know that $\ln g_n(T|X)-\hat{l}_n(T|X) \xrightarrow{P} 0$ and from Condition 3, we have $\hat{l}_n(T|X)-l(T) \xrightarrow{P} 0.$ Adding this up using the fact that convergence in probability is preserved under addition, we know that $\ln g_n(T|X) \xrightarrow{P} l(T)$.

Hence by taking the limit of (\ref{thmeqn1}) as $n$ grows, we have that $$\lim_{n \rightarrow \infty} l(T_n|X) \geq l(T^*_{\textrm{MLE}}).$$ Since $T^*_{\textrm{MLE}}$ is optimal for $l(T)$ by definition, and by Condition 4, we conclude that $T_n$ stays in $[T^*_{\textrm{MLE}}]$ when $n$ is sufficiently large. \qed

\section{Conclusion}\label{sec:conclusion}
In this work, we have presented a Bayesian approach to density estimation using sparse piece-wise constant estimators for categorical (or binary) data. 
Our methods have nice properties, including that their prior encourages sparsity, which permits interpretability. 

For tree-based methods, the prior is the user's desired number of leaves or branches in each node in the tree, while for the density rule list, the prior regularizes the length of the list. We designed a simulated annealing scheme, which alongside the inclusion-exclusion principle, and efficient data representation via bit vectors, allows us to find an optimal solution relatively fast.

Our methods outperform existing baselines. They do not have the pitfalls of other nonparametric density estimation methods like density estimation trees, which are top-down greedy. Further, they are consistent, without needing to asymptotically produce infinitesimally small leaves. 

The interpretability of density trees and rule lists allows easier visualization of the estimated density values, aiding in the understanding of the data distribution, detection of outliers and errors, model selection, and could assist with decision-making.
The approaches presented here have given us insight into real data sets (including the housebreak data set from the Cambridge Police and COCO-stuff labels), that we could not have obtained reliably in any other way.


\bibliographystyle{plainnat}
\bibliography{macform16}

\begin{thebibliography}{57}
\providecommand{\natexlab}[1]{#1}
\providecommand{\url}[1]{\texttt{#1}}
\expandafter\ifx\csname urlstyle\endcsname\relax
  \providecommand{\doi}[1]{doi: #1}\else
  \providecommand{\doi}{doi: \begingroup \urlstyle{rm}\Url}\fi

\bibitem[Abou-Jaoude(1976)]{abou1976conditions}
Saab Abou-Jaoude.
\newblock Conditions n{\'e}cessaires et suffisantes de convergence l1 en probabilit{\'e} de l'histogramme pour une densit{\'e}.
\newblock \emph{Annales de l'IHP Probabilit{\'e}s et statistiques}, 12\penalty0 (3):\penalty0 213--231, 1976.

\bibitem[Akaike(1954)]{akaike1954approximation}
Hirotugu Akaike.
\newblock An approximation to the density function.
\newblock \emph{Annals of the Institute of Statistical Mathematics}, 6\penalty0 (2):\penalty0 127--132, 1954.

\bibitem[Anderlini(2015)]{anderlini2015density}
Lucio Anderlini.
\newblock Density estimation trees in high energy physics.
\newblock \emph{arXiv preprint arXiv:1502.00932}, 2015.

\bibitem[Anderlini(2016)]{anderlini2016density}
Lucio Anderlini.
\newblock Density estimation trees as fast non-parametric modelling tools.
\newblock \emph{Journal of Physics: Conference Series}, 762\penalty0 (1):\penalty0 012042, 2016.

\bibitem[Angelino et~al.(2018)Angelino, Larus-Stone, Alabi, Seltzer, and Rudin]{AngelinoEtAl2018}
Elaine Angelino, Nicholas Larus-Stone, Daniel Alabi, Margo Seltzer, and Cynthia Rudin.
\newblock Learning certifiably optimal rule lists for categorical data.
\newblock \emph{Journal of Machine Learning Research}, 18\penalty0 (234):\penalty0 1--78, 2018.

\bibitem[Cacoullos(1966)]{cacoullos1966estimation}
Theophilos Cacoullos.
\newblock Estimation of a multivariate density.
\newblock \emph{Annals of the Institute of Statistical Mathematics}, 18\penalty0 (1):\penalty0 179--189, 1966.

\bibitem[Caesar et~al.(2018)Caesar, Uijlings, and Ferrari]{caesar2018cvpr}
Holger Caesar, Jasper Uijlings, and Vittorio Ferrari.
\newblock {COCO-Stuff}: Thing and stuff classes in context.
\newblock In \emph{Computer vision and pattern recognition (CVPR), 2018 IEEE conference on}. IEEE, 2018.

\bibitem[Cattaneo et~al.(2019)Cattaneo, Jansson, and Ma]{cattaneo2019simple}
Matias~D Cattaneo, Michael Jansson, and Xinwei Ma.
\newblock Simple local polynomial density estimators.
\newblock \emph{Journal of the American Statistical Association}, pages 1--7, 2019.

\bibitem[Chakrabarti et~al.(2006)Chakrabarti, Ester, Fayyad, Gehrke, Han, Morishita, Piatetsky-Shapiro, and Wang]{chakrabarti2006data}
Soumen Chakrabarti, Martin Ester, Usama Fayyad, Johannes Gehrke, Jiawei Han, Shinichi Morishita, Gregory Piatetsky-Shapiro, and Wei Wang.
\newblock Data mining curriculum: A proposal (version 1.0).
\newblock \emph{Intensive Working Group of ACM SIGKDD Curriculum Committee}, 2006.

\bibitem[Chen et~al.(2006)Chen, Morris, and Martin]{chen2006probability}
Tao Chen, Julian Morris, and Elaine Martin.
\newblock Probability density estimation via an infinite gaussian mixture model: application to statistical process monitoring.
\newblock \emph{Journal of the Royal Statistical Society: Series C (Applied Statistics)}, 55\penalty0 (5):\penalty0 699--715, 2006.

\bibitem[Chipman et~al.(2010)Chipman, George, and McCulloch]{chipman2010bart}
Hugh~A Chipman, Edward~I George, and Robert~E McCulloch.
\newblock Bart: Bayesian additive regression trees.
\newblock \emph{The Annals of Applied Statistics}, 4\penalty0 (1):\penalty0 266--298, 2010.

\bibitem[Devroye et~al.(1996)Devroye, Gy{\"{o}}rfi, and Lugosi]{citeulike:3463149}
L.~Devroye, L.~Gy{\"{o}}rfi, and G.~Lugosi.
\newblock \emph{{A Probabilistic Theory of Pattern Recognition}}.
\newblock Springer, 1996.

\bibitem[Devroye(1991)]{devroye1991exponential}
Luc Devroye.
\newblock Exponential inequalities in nonparametric estimation.
\newblock In \emph{Nonparametric functional estimation and related topics}, pages 31--44. Springer, 1991.

\bibitem[Devroye and Gy{\"o}rfi(1983)]{devroye1983distribution}
Luc Devroye and L{\'a}szl{\'o} Gy{\"o}rfi.
\newblock Distribution-free exponential bound on the l1 error of partitioning estimates of a regression function.
\newblock \emph{Probability and statistical decision theory, Vol. A}, 67:\penalty0 76, 1983.

\bibitem[Fisher et~al.(2019)Fisher, Rudin, and Dominici]{FisherRuDo19}
Aaron Fisher, Cynthia Rudin, and Francesca Dominici.
\newblock All models are wrong, but many are useful: Learning a variable's importance by studying an entire class of prediction models simultaneously.
\newblock \emph{Journal of Machine Learning Research}, 20\penalty0 (177):\penalty0 1--81, 2019.
\newblock URL \url{http://jmlr.org/papers/v20/18-760.html}.

\bibitem[Friedman et~al.(1984)Friedman, Stuetzle, and Schroeder]{friedman1984projection}
Jerome~H Friedman, Werner Stuetzle, and Anne Schroeder.
\newblock Projection pursuit density estimation.
\newblock \emph{Journal of the American Statistical Association}, 79\penalty0 (387):\penalty0 599--608, 1984.

\bibitem[Holmstr{\"o}m et~al.(2015)Holmstr{\"o}m, Karttunen, and Klemel{\"a}]{holmstrom2015estimation}
Lasse Holmstr{\"o}m, Ky{\"o}sti Karttunen, and Jussi Klemel{\"a}.
\newblock Estimation of level set trees using adaptive partitions.
\newblock \emph{Computational Statistics}, 32\penalty0 (3):\penalty0 1139--1163, 2015.

\bibitem[Jebara(2008)]{jebara2012bayesian}
Tony~S Jebara.
\newblock Bayesian out-trees.
\newblock In \emph{Proceedings of the Twenty-Fourth Conference on Uncertainty in Artificial Intelligence {(UAI)}}, 2008.

\bibitem[Letham et~al.(2015)Letham, Rudin, McCormick, and Madigan]{LethamRuMcMa15}
Benjamin Letham, Cynthia Rudin, Tyler~H. McCormick, and David Madigan.
\newblock Interpretable classifiers using rules and bayesian analysis: Building a better stroke prediction model.
\newblock \emph{Annals of Applied Statistics}, 9\penalty0 (3):\penalty0 1350--1371, 2015.

\bibitem[Li et~al.(2016)Li, Yang, and Wong]{yang2015density}
Dangna Li, Kun Yang, and Wing~Hung Wong.
\newblock Density estimation via discrepancy based adaptive sequential partition.
\newblock In D.~D. Lee, M.~Sugiyama, U.~V. Luxburg, I.~Guyon, and R.~Garnett, editors, \emph{Advances in Neural Information Processing Systems 29}, pages 1091--1099. Curran Associates, Inc., 2016.

\bibitem[Li and Barron(2000)]{NIPS1999_1673}
Jonathan~Q. Li and Andrew~R. Barron.
\newblock Mixture density estimation.
\newblock In S.~A. Solla, T.~K. Leen, and K.~M\"{u}ller, editors, \emph{Advances in Neural Information Processing Systems 12}, pages 279--285. MIT Press, {Denver, CO}, 2000.

\bibitem[Liu et~al.(2007)Liu, Lafferty, and Wasserman]{AISTATS07_LiuLW}
Han Liu, John Lafferty, and Larry Wasserman.
\newblock Sparse nonparametric density estimation in high dimensions using the rodeo.
\newblock In Marina Meila and Xiaotong Shen, editors, \emph{Proceedings of the Eleventh International Conference on Artificial Intelligence and Statistics}, volume~2 of \emph{Proceedings of Machine Learning Research}, pages 283--290, San Juan, Puerto Rico, 21--24 Mar 2007. PMLR.

\bibitem[Liu et~al.(2011)Liu, Xu, Gu, Gupta, Lafferty, and Wasserman]{Liu:2011:FDE:1953048.2021032}
Han Liu, Min Xu, Haijie Gu, Anupam Gupta, John Lafferty, and Larry Wasserman.
\newblock Forest density estimation.
\newblock \emph{Journal of Machine Learning Research}, 12:\penalty0 907--951, July 2011.
\newblock ISSN 1532-4435.

\bibitem[Liu et~al.(2021)Liu, Xu, Jiang, and Wong]{Liue2101344118}
Qiao Liu, Jiaze Xu, Rui Jiang, and Wing~Hung Wong.
\newblock Density estimation using deep generative neural networks.
\newblock \emph{Proceedings of the National Academy of Sciences}, 118\penalty0 (15), 2021.
\newblock ISSN 0027-8424.
\newblock \doi{10.1073/pnas.2101344118}.
\newblock URL \url{https://www.pnas.org/content/118/15/e2101344118}.

\bibitem[Lu et~al.(2013)Lu, Jiang, and Wong]{lu2013multivariate}
Luo Lu, Hui Jiang, and Wing~H Wong.
\newblock Multivariate density estimation by {B}ayesian sequential partitioning.
\newblock \emph{Journal of the American Statistical Association}, 108\penalty0 (504):\penalty0 1402--1410, 2013.

\bibitem[Lugosi and Nobel(1996)]{lugosi1996consistency}
G{\'a}bor Lugosi and Andrew Nobel.
\newblock Consistency of data-driven histogram methods for density estimation and classification.
\newblock \emph{The Annals of Statistics}, 24\penalty0 (2):\penalty0 687--706, 1996.

\bibitem[Luo et~al.(2019)Luo, Liu, and Wang]{luo2019combining}
Runfei Luo, Anna Liu, and Yuedong Wang.
\newblock Combining smoothing spline with conditional gaussian graphical model for density and graph estimation.
\newblock \emph{arXiv preprint arXiv:1904.00204}, 2019.

\bibitem[Mahapatruni and Gray(2011)]{mahapatruni2011cake}
Ravi Sastry~Ganti Mahapatruni and Alexander Gray.
\newblock Cake: Convex adaptive kernel density estimation.
\newblock In Geoffrey Gordon, David Dunson, and Miroslav Dudík, editors, \emph{Proceedings of the Fourteenth International Conference on Artificial Intelligence and Statistics}, volume~15 of \emph{Proceedings of Machine Learning Research}, pages 498--506, Fort Lauderdale, FL, USA, 11--13 Apr 2011. PMLR.

\bibitem[M{\"u}ller and Quintana(2004)]{muller2004nonparametric}
Peter M{\"u}ller and Fernando~A Quintana.
\newblock Nonparametric {B}ayesian data analysis.
\newblock \emph{Statistical Science}, 19\penalty0 (1):\penalty0 95--110, 2004.

\bibitem[Nadaraya(1970)]{nadaraya1970remarks}
{\'E}~A Nadaraya.
\newblock Remarks on non-parametric estimates for density functions and regression curves.
\newblock \emph{Theory of Probability \& Its Applications}, 15\penalty0 (1):\penalty0 134--137, 1970.

\bibitem[Ooi(2012)]{ooi2012density}
Hong Ooi.
\newblock Density visualization and mode hunting using trees.
\newblock \emph{Journal of Computational and Graphical Statistics}, 11\penalty0 (2):\penalty0 328--347, 2012.

\bibitem[Ormoneit and Tresp(1996)]{ormoneit1995improved}
Dirk Ormoneit and Volker Tresp.
\newblock Improved gaussian mixture density estimates using bayesian penalty terms and network averaging.
\newblock In D.~S. Touretzky, M.~C. Mozer, and M.~E. Hasselmo, editors, \emph{Advances in Neural Information Processing Systems 8}, pages 542--548. MIT Press, Denver, USA, 1996.

\bibitem[Ormoneit and Tresp(1998)]{ormoneit1998averaging}
Dirk Ormoneit and Volker Tresp.
\newblock Averaging, maximum penalized likelihood and bayesian estimation for improving gaussian mixture probability density estimates.
\newblock \emph{IEEE Transactions on Neural Networks}, 9\penalty0 (4):\penalty0 639--650, 1998.

\bibitem[Parzen(1962)]{parzen1962estimation}
Emanuel Parzen.
\newblock On estimation of a probability density function and mode.
\newblock \emph{The Annals of Mathematical Statistics}, 33\penalty0 (3):\penalty0 1065--1076, 1962.

\bibitem[Patki et~al.(2016)Patki, Wedge, and Veeramachaneni]{patki2016synthetic}
Neha Patki, Roy Wedge, and Kalyan Veeramachaneni.
\newblock The synthetic data vault.
\newblock In \emph{2016 {IEEE} International Conference on Data Science and Advanced Analytics ({DSAA})}, pages 399--410. IEEE, 2016.

\bibitem[Ram and Gray(2011)]{Ram:2011:DET:2020408.2020507}
Parikshit Ram and Alexander~G Gray.
\newblock Density estimation trees.
\newblock In \emph{Proceedings of the 17th ACM SIGKDD international conference on Knowledge discovery and data mining}, pages 627--635, 2011.

\bibitem[Rehn et~al.(2018)Rehn, Ahmadi, and Kramer]{rehn2018forest}
Patrick Rehn, Zahra Ahmadi, and Stefan Kramer.
\newblock Forest of normalized trees: Fast and accurate density estimation of streaming data.
\newblock In \emph{2018 IEEE 5th International Conference on Data Science and Advanced Analytics (DSAA)}, pages 199--208. IEEE, 2018.

\bibitem[Rejt{\"o} and R{\'e}v{\'e}sz(1973)]{rejto1973density}
L~Rejt{\"o} and P~R{\'e}v{\'e}sz.
\newblock Density estimation and pattern classification.
\newblock \emph{Problems of Control and Information Theory}, 2\penalty0 (1):\penalty0 67--80, 1973.

\bibitem[Rosenblatt et~al.(1956)]{rosenblatt1956remarks}
Murray Rosenblatt et~al.
\newblock Remarks on some nonparametric estimates of a density function.
\newblock \emph{The Annals of Mathematical Statistics}, 27\penalty0 (3):\penalty0 832--837, 1956.

\bibitem[Rudin and Ertekin(2018)]{ErtekinRu18}
Cynthia Rudin and Seyda Ertekin.
\newblock Learning customized and optimized lists of rules with mathematical programming.
\newblock \emph{Mathematical Programming C (Computation)}, 10:\penalty0 659--702, 2018.

\bibitem[Sasaki and Hyv{\"a}rinen(2018)]{sasaki2018neural}
Hiroaki Sasaki and Aapo Hyv{\"a}rinen.
\newblock Neural-kernelized conditional density estimation.
\newblock \emph{arXiv preprint arXiv:1806.01754}, 2018.

\bibitem[Scott(1979)]{scott1979optimal}
David~W Scott.
\newblock On optimal and data-based histograms.
\newblock \emph{Biometrika}, 66\penalty0 (3):\penalty0 605--610, 1979.

\bibitem[Seidl et~al.(2009)Seidl, Assent, Kranen, Krieger, and Herrmann]{seidl2009indexing}
Thomas Seidl, Ira Assent, Philipp Kranen, Ralph Krieger, and Jennifer Herrmann.
\newblock Indexing density models for incremental learning and anytime classification on data streams.
\newblock In \emph{In 12th EDBT/ICDT}, pages 311--322, Saint Petersburg, Russia, 2009. ACM.

\bibitem[Semenova et~al.(2022)Semenova, Rudin, and Parr]{SemenovaRuPa2022}
Lesia Semenova, Cynthia Rudin, and Ronald Parr.
\newblock On the existence of simpler machine learning models.
\newblock In \emph{{ACM} Conference on Fairness, Accountability, and Transparency ({ACM FAccT})}, 2022.

\bibitem[Silverman(1986)]{silverman1986density}
Bernard~W Silverman.
\newblock \emph{Density estimation for statistics and data analysis}, volume~26.
\newblock CRC press, 1986.

\bibitem[Varet et~al.(2023)Varet, Lacour, Massart, and Rivoirard]{varet2019numerical}
Suzanne Varet, Claire Lacour, Pascal Massart, and Vincent Rivoirard.
\newblock Numerical performance of penalized comparison to overfitting for multivariate kernel density estimation.
\newblock \emph{ESAIM: Probability and Statistics}, 27:\penalty0 621--667, 2023.

\bibitem[Wand(1997)]{wand1997data}
MP~Wand.
\newblock Data-based choice of histogram bin width.
\newblock \emph{The American Statistician}, 51\penalty0 (1):\penalty0 59--64, 1997.

\bibitem[Wasserman(2006)]{wasserman2006all}
Larry Wasserman.
\newblock \emph{All of nonparametric statistics}.
\newblock Springer Science \& Business Media, 2006.

\bibitem[Willett and Nowak(2007)]{willett2007minimax}
RM~Willett and Robert~D Nowak.
\newblock Minimax optimal level-set estimation.
\newblock \emph{IEEE Transactions on Image Processing}, 16\penalty0 (12):\penalty0 2965--2979, 2007.

\bibitem[Wong and Ma(2010)]{wong2010optional}
Wing~H Wong and Li~Ma.
\newblock Optional p{\'o}lya tree and {B}ayesian inference.
\newblock \emph{The Annals of Statistics}, 38\penalty0 (3):\penalty0 1433--1459, 2010.

\bibitem[Wu et~al.(2018)Wu, Hou, and Yang]{wu2018density}
Kaiyuan Wu, Wei Hou, and Hongbo Yang.
\newblock Density estimation via the random forest method.
\newblock \emph{Communications in Statistics-Theory and Methods}, 47\penalty0 (4):\penalty0 877--889, 2018.

\bibitem[Wu et~al.(2007)Wu, Tjelmeland, and West]{wu2007bayesian}
Yuhong Wu, H{\aa}kon Tjelmeland, and Mike West.
\newblock Bayesian {CART}: Prior specification and posterior simulation.
\newblock \emph{Journal of Computational and Graphical Statistics}, 16\penalty0 (1):\penalty0 44--66, 2007.

\bibitem[Yang et~al.(2017)Yang, Rudin, and Seltzer]{YangRuSe16}
Hongyu Yang, Cynthia Rudin, and Margo Seltzer.
\newblock Scalable {B}ayesian rule lists.
\newblock In \emph{Proceedings of the 34th International Conference on Machine Learning {(ICML)}}, 2017.

\bibitem[Yang and Wong(2014{\natexlab{a}})]{yang2014density}
Kun Yang and Wing~Hung Wong.
\newblock Density estimation via adaptive partition and discrepancy control.
\newblock \emph{arXiv preprint arXiv:1404.1425}, 2014{\natexlab{a}}.

\bibitem[Yang and Wong(2014{\natexlab{b}})]{yang2014discovering}
Kun Yang and Wing~Hung Wong.
\newblock Discovering and visualizing hierarchy in the data.
\newblock \emph{arXiv preprint arXiv:1403.4370}, 2014{\natexlab{b}}.

\bibitem[Zhao et~al.(1991)Zhao, Krishnaiah, and Chen]{zhao1991almost}
Lin~Cheng Zhao, Paruchuri~R Krishnaiah, and Xi~Ru Chen.
\newblock Almost sure $\ell_r$-norm convergence for data-based histogram density estimates.
\newblock \emph{Theory of Probability \& Its Applications}, 35\penalty0 (2):\penalty0 396--403, 1991.

\bibitem[Zhuang et~al.(1996)Zhuang, Huang, Palaniappan, and Zhao]{zhuang1996gaussian}
Xinhua Zhuang, Yan Huang, Kannappan Palaniappan, and Yunxin Zhao.
\newblock Gaussian mixture density modeling, decomposition, and applications.
\newblock \emph{IEEE Transactions on Image Processing}, 5\penalty0 (9):\penalty0 1293--1302, 1996.

\end{thebibliography}








\begin{APPENDICES}
\section{Optimal Density for the Likelihood Function}\label{appendix:optimal_density}

Denote the pointwise density estimate at $x$ to be $\hat{f}_{n,x}=\frac{n_{x}}{n}$. Denote the density estimate for all points within leaf $l$ similarly as $\hat{f}_{n,l}=\frac{\sum_{j\in \textrm{leaf }l} n_j}{n V_{l}}$.


The true density from which the data are assumed to be generated is denoted $D$. We assume that $D$ arises from a tree over the input space.

\noindent \textbf{Lemma 1}: Any tree achieving the maximum likelihood on the training data has pointwise density equal to $\hat{f_n}(x)$. This means for any $l$ in the tree and for any $x$, $\hat{f}_{n,l}(x)=\hat{f}_{n,x}(x)$.


\noindent\textbf{Proof:}

We would like to show that points should be grouped together if and only if they share the same density. Clearly, if points have the same density, grouping them together will preserve constant density. Thus, the backwards implication holds. We have only to show that if points do not share the same density, we should not group them together.

We will show that the pointwise histogram becomes better than using the tree if the tree is not correct. This is a variation of a well-known result that the maximum likelihood is the pointwise maximum likelihood.
We will show:
\begin{equation*} 
\prod_{j\in \textrm{leaf } l} \hat{f}_{n,j}^{n_j} \geq \hat{f}_{n,l}^{\sum_{j\in \textrm{leaf } l}  n_j}. 
\end{equation*}
By taking logarithms, this reduces to
\begin{align}
\sum_{j\in \textrm{leaf } l} n_j \log \hat{f}_{n,j} &\geq \sum_{j\in \textrm{leaf } l} n_j \log \hat{f}_{n,l} \label{aftertakinglog}\\
\sum_{j\in \textrm{leaf } l} n_j \log \left( \frac{n_j}{n} \right) &\geq \sum_{j\in \textrm{leaf } l} n_j \log \left( \frac{\sum_{m\in \textrm{leaf } l  } n_m}{n V_l}\right) \nonumber\\
\sum_{j\in \textrm{leaf } l} n_j \log n_j &\geq  \sum_{j\in \textrm{leaf } l} n_j \log \left( \frac{\sum_{m\in \textrm{leaf } l} n_m}{ V_l}\right) \nonumber \\
\sum_ {j\in \textrm{leaf } l} n_j\log \left( \frac{n_j}{\sum_{m\in \textrm{leaf } l } n_m} \right) \nonumber  &\geq \sum_{m\in \textrm{leaf } l} n_j \log \left( \frac{1}{ \sum_{m\in \textrm{leaf } l} V_m} \right) \nonumber \\
\sum_{j\in \textrm{leaf } l} \frac{n_j}{\sum_{m\in \textrm{leaf } l }  n_m} \log \left( \frac{n_j}{\sum_{m\in \textrm{leaf } l } n_m}\right) 
&\geq \sum_{j\in \textrm{leaf } l }  \frac{n_j}{\sum_{m\in \textrm{leaf } l} n_m}\log \left( \frac{1}{ \sum_{m\in \textrm{leaf } l } V_m}\right).\nonumber
\end{align}
The last equation follows from Gibb's inequality. Hence we have proven that the statement is true. 

To avoid singularities, we separately consider the case when one of the $\hat{f}_{n,j}=0$. For a particular value of $q$, if $\hat{f}_{n,q}=0$, then $n_q=0$ by definition. Hence, if we include a new $x$ within the leaf that has no training examples, we will find that the left hand side term of (\ref{aftertakinglog}) remains the same but since the volume increases when we add the new point, the quantity on the right decreases. Hence the inequality still holds.

\section{Discussion on Implementation of DET}\label{appendix:det_implementation}

 \begin{algorithm}[t]
      \textbf{Input}:  categorical feature $f_c$, $c_1,...,c_J$ - categories of the feature $f_c$ \\
    \textbf{Output}: {real-valued feature $f_r$}

\begin{algorithmic}[1]
    \FOR{every category $c_j$}
    \STATE compute its frequency $p_j$ in the data set
    \ENDFOR
    \STATE sort the categories from the most frequent to the least frequent
    \STATE split interval $[0,1]$ in $J$ frequency intervals $I_j = [a_j, b_j]$ based on the cumulative probability, meaning that the length of the interval with index $j$ is $p_j$ 
    \FOR{every sample $i$}
    \STATE find its category $c_j^i$
    \STATE find corresponding frequency interval $I_j = [a_j, b_j]$
    \STATE chose a value $v \in I_j$ by sampling from a truncated Gaussian distribution with  $\mu = a_j + p_j / 2$ and $\sigma = (b_j - a_j) / 6$
    \STATE assign $v$ to $f_r^i$
    \ENDFOR
\caption{Conversion of categorical feature to real}\label{algo:sampling}
\end{algorithmic}
\end{algorithm}

 DET implemented in Python from MLPACK library \citep{Ram:2011:DET:2020408.2020507} is designed for continuous variables. In order to compare our methods to DET, we  pre-processed data sets using one-hot encoding and the Synthetic Data Vault algorithm from \cite{patki2016synthetic} (see Algorithm \ref{algo:sampling}). For the one-hot encoding, we create dummy variables for every category and then we called DET on the data set. The result included probability densities that sum to values much larger than one, which is not a valid result. Specifically, the DET code returned density values for a single observation as large as $12.71$ for titanic data (when the sum of density values should be 1). Such values should not be returned if the code was designed for categorical data, because density is equal to the probability divided by the volume, and the volume for each categorical bin is at least $1$; the density value in each bin must be at most $1$ in order to return valid density results. Given such high density values, the log-likelihood was positive, for example in the range [396, 13107] for crime data for different hyperparameter settings (the correct values are always negative). Utilizing other pre-processing methods, such as the Synthetic Data Vault algorithm, decreased values of the log-likelihood, but the problem of densities larger than 1 for single observations was still present. 

 Because of the issues in the density computation (and thus likelihood computation) discussed above, we needed to figure out a way to compare performance of DET with other methods. In order to compare the performance, we used a one-hot encoding representation of the data, and used the DET algorithm to create leaves. Then, we utilized our own objective function to compute the log-likelihood from Method I and II. Trees that were returned by DET had splits based on one category in the nodes (due to the nature of encoding and the DET algorithm), which resulted in more leaves on average and lower log-likelihood as compared to our optimized methods.
 
 
 If, in the future, one created an implementation of DET that produces valid probability densities for categorical data, our objective functions can be used as criteria to select a suitable density tree. This algorithm would still have the disadvantage of producing greedy trees rather than optimized trees.

\section{Discussion on Run time Performance}\label{appendix:runtime}

\begin{table}[ht]
    \centering
    \footnotesize
    \begin{tabular}{|p{0.17\textwidth}|p{0.07\linewidth}|p{0.07\textwidth}||p{0.085\textwidth}|p{0.085\textwidth}|p{0.085\textwidth}||p{0.085\textwidth}|p{0.085\textwidth}|p{0.085\textwidth}|}
    \hline
        Data Set 	&	 Number of train points 	&	 Number of feature-value pairs 	&	 Leaf-sparse, best, (sec)  	&	 Branch-sparse, best, (sec) 	&	 Rule list, best, (sec) 	&	 Leaf-sparse, multiple, (sec) 	&	  Branch-sparse, multiple, (sec)  	&	 Rule list, multiple, (sec) \\ \hline \hline	
        Balance  	&	531	&	20	&	\textbf{0.207}	&	1.036	&	8.226	&	0.587	&	2.166	&	82.402	\\\hline
        Bank Full 	&	38429	&	51	&	\textbf{19.652}	&	21.115	&	31.340	&	61.567	&	42.971	&	189.415	\\\hline
        Car 	&	1468	&	21	&	\textbf{0.276}	&	1.324	&	7.123	&	0.826	&	2.610	&	88.097	\\\hline
        Chess (King-Rook vs. King-Pawn) 	&	2716	&	73	&	\textbf{2.808}	&	3.631	&	58.269	&	8.683	&	7.147	&	246.194	\\\hline
        COCO stuff+thing labels hierarchy 	&	1607458	&	206	&	\textbf{315.418}	&	339.948	&	788.144	&	886.045	&	632.347	&	1175.841	\\\hline
        COCO staff labels 	&	118280	&	182	&	\textbf{119.142}	&	169.533	&	1056.583	&	382.756	&	316.800	&	1999.329	\\\hline
        COCO things labels	&	117266	&	160	&	\textbf{104.491}	&	127.561	&	545.789	&	278.445	&	247.029	&	1476.896	\\\hline
        COMPAS 	&	6489	&	19	&	\textbf{1.225}	&	2.171	&	14.503	&	3.606	&	4.345	&	95.327	\\\hline
        Connect 4 	&	57423	&	126	&	\textbf{53.692}	&	58.286	&	404.347	&	143.833	&	111.995	&	882.699	\\\hline
        Crime 	&	3178	&	24	&	\textbf{1.393}	&	1.998	&	24.709	&	3.990	&	3.933	&	120.641	\\\hline
        HELOC 	&	8890	&	195	&	\textbf{19.402}	&	23.349	&	111.346	&	62.415	&	48.889	&	607.118	\\\hline
        Mushroom 	&	4797	&	100	&	\textbf{5.317}	&	6.362	&	31.890	&	16.360	&	12.685	&	699.592	\\\hline
        Nursery 	&	11016	&	27	&	\textbf{1.196}	&	4.418	&	10.667	&	3.587	&	8.571	&	106.414	\\\hline
         Telco Customer Churn 	&	5977	&	73	&	\textbf{6.405}	&	7.402	&	138.462	&	17.196	&	14.860	&	370.096	\\\hline
        Titanic 	&	1761	&	8	&	\textbf{0.324}	&	0.425	&	7.323	&	0.947	&	0.804	&	41.436	\\\hline
        US Census 1990 (1m) 	&	1000000	&	396	&	3580.249	&	\textbf{2813.563}	&	8382.199	&	10602.726	&	6299.193	&	10145.797	\\\hline
        US Census 1990 	&	2089542	&	396	&	\textbf{6216.533}	&	8965.318	&	31199.718	&	19469.654	&	16118.220	&	39081.668	\\\hline
    \end{tabular}
    \caption{Run time analysis of three methods for different complexity data sets. All time measurements are averaged over 5 repeats. ``Leaf-sparse, best,'' ``Branch-sparse, best,'' and ``Rule list, best'' show the average run time in seconds of the methods with best parameters $\lambda$ and $\eta$. ``Leaf-sparse, multiple,'' ``Branch-sparse, multiple,'' and ``Rule list, multiple'' show the average run time in seconds of the methods, including the time needed to try multiple parameters.}
    \label{table:runtime_results}
\end{table}

We measured the performance of density estimation methods on 17 categorical data sets that are described in Table \ref{table:datasets_description}. Continuous features in HELOC and Telco Customer Churn data sets were divided into 10 bins uniformly to create categorical features. We considered labels from the COCO stuff+thing data set, and this resulted in three data sets: (1) COCO stuff that consists of binary features, where each indicates whether or not the stuff category is present in the image; (2) COCO thing that is built the same way except features are thing categories; (3) COCO stuff+thing, a four feature data set that utilizes the hierarchical structure of COCO detection labels, where each feature is a hierarchy level (such as ``animal,'' ``dog,'' ``things,'' or ``outdoor'' where a ``dog'' is an ``animal'' and is ``outdoor'' and is in the ``thing'' data set). We removed data samples with missing values. The train and validation data split for the run time experiments is fixed and shown in Table \ref{table:datasets_description}.

For each setting of the parameters for each algorithm, we ran the algorithm five times to account for randomness in the optimization. We chose the best parameter values, and reported the average run time (over the 5 repeats) for these best parameters. We also reported the average (over the 5 repeats) total run time, including the time needed to choose parameter values.  
For the leaf-sparse density tree model, parameter $\lambda$ (number of leaves) was chosen from the set ${5, 8, 10}$. For the branch-sparse density tree,  $\lambda$ (number of branches) was chosen from ${2, 3}$; for the sparse density rule list $\lambda$ (length of the list), was chosen from the set ${3, 5, 7}$; and $\eta$ (number of conjunctions in a rule) was chosen from ${1, 2}$. $\alpha$ was fixed to be 2 for the tree-based methods and 1 for the density rule list. 
Run time results for all 17 data sets are shown in Table \ref{table:runtime_results}. For tree-based methods, the run time for evaluating multiple parameters is approximately three times (for leaf-sparse) and two times (for branch-sparse) larger than the run time for the methods when we knew the best parameters, simply because each run of the algorithm took approximately the same amount of time. For the density rule list, a significant portion of the run time is spent on data and volume pre-processing computations that are executed only once at the beginning. Thus, running the algorithm with multiple (i.e., 6) parameters is not 6 times the run time for running once when knowing the best parameters.

Run times in Table \ref{table:runtime_results} are computed by running our methods on the Duke University’s Computer Science Department cluster. On a single CPU machine (Intel(R) Core(TM) i7-6700 CPU @ 3.40GHz), for the Titanic dataset the average run
time (over 5 iterations) for the leaf-based method is 0.276 sec, branch-based -- 0.416 sec, density list -- 6.928 sec; for Crime dataset: leaf-based -- 0.1 sec, branch-based -- 1.4 sec, density list -- 12.28 sec; for Balance dataset: leaf-based -- 0.202 sec, branch-based -- 0.99 sec, density list -- 7.544 sec; Bank Full dataset: leaf-based -- 20.218 sec, branch-based -- 23.077 sec, density list -- 23.722 sec. The run times are reported for one average run of the algorithms.

\begin{table}[!ht]
    \centering
    \footnotesize
    \begin{tabular}{|p{0.17\textwidth}|p{0.075\textwidth}|p{0.07\textwidth}|p{0.07\textwidth}|p{0.06\textwidth}|p{0.07\textwidth}|p{0.08\textwidth}|p{0.24\textwidth}|}
    \hline
        Data Set & Total number of points & Number of train points & Number of validation points & Train validation split & Number of features & Number of feature-value pairs & Processing notes \\ \hline
        Balance  & 625 & 531 & 94 & 15\% & 4 & 20 & ~ \\ \hline
        Bank Full & 45211 & 38429 & 6782 & 15\% & 12 & 51 & ~ \\ \hline
        Car & 1728 & 1468 & 260 & 15\% & 6 & 21 & ~ \\ \hline
        Chess (King-Rook vs. King-Pawn) & 3196 & 2716 & 480 & 15\% & 36 & 73 & ~ \\ \hline
        COCO stuff+thing labels hierarchy & 1677309 & 1607458 & 69582 & 4\% & 4 & 206 & Features are formed from COCO detection labels hierarchy: stuff or thing, indoor or outdoor, super-category, and category \\ \hline
        COCO stuff labels & 123280 & 118280 & 5000 & 4\% & 91 & 182 & Features are detection label categories \\ \hline
        COCO thing labels & 122218 & 117266 & 4952 & 4\% & 80 & 160 & Features are detection label categories \\ \hline
        COMPAS & 7210 & 6489 & 721 & 10\% & 7 & 19 & ~ \\ \hline
        Connect 4 & 67557 & 57423 & 10134 & 15\% & 42 & 126 & ~ \\ \hline
        Crime & 3739 & 3178 & 561 & 15\% & 6 & 24 & ~ \\ \hline
        HELOC & 10459 & 8890 & 1569 & 15\% & 23 & 195 & Cut all features in 10 bins \\ \hline
        Mushroom & 5644 & 4797 & 847 & 15\% & 23 & 100 & Dropped entries with missing values \\ \hline
        Nursery & 12960 & 11016 & 1944 & 15\% & 8 & 27 & ~ \\ \hline
        Telco Customer Churn  & 7032 & 5977 & 1055 & 15\% & 19 & 73 & Cut 3 continuous features in 10 bins; Dropped entries with missing values \\ \hline
        Titanic & 2201 & 1761 & 440 & 20\% & 3 & 8 & ~ \\ \hline
        US Census 1990 (1m) & 1150000 & 1000000 & 150000 & 15\% & 68 & 396 & Considered 1 million samples \\ \hline
        US Census 1990 & 2458285& 2089542 & 368743 & 13\% & 68 & 396 & ~ \\ \hline
    \end{tabular}
    \caption{Data sets statistics, including details on data set size, number of features and categories, and pre-processing notes if any.}
    \label{table:datasets_description}
\end{table}

\section{Recommendations on the Choice of Algorithm and Parameters}\label{appendix:user_guide}

From the experiments we conducted, we found that the leaf-based density trees were the most useful and intuitive. They also run the fastest (Table \ref{table:runtime_results}) for the vast majority of the datasets we considered.

 \textbf{Leaf-based vs$.$ branch-based.} Leaf-based and branch-based methods are similar in structure but differ in how the shape of the model is controlled by the prior, specifically, the user controls either the number of leaves or the number of branches. 
 This matters when there are many categories per feature: the leaf-based approach may try to put these categories in one node, while the branch-based method might keep them in separate nodes. However, it takes longer to run the branch-based method (Table \ref{table:runtime_results}), and its models are typically more complex than the leaf-based method's models on the same dataset (Table \ref{tableS:best_params}).

\textbf{Trees vs Lists.} When comparing density trees and density rule lists, rule lists are one-sided trees, but they have multiple conditions defining each rule. Density rule lists can be more helpful if the user prefers a very sparse density model or has a smaller dataset. Concerning run time, the rule lists were the slowest method per run on average. However, one of the major bottlenecks for rule lists was memory space and time needed to process the data and mine the rules. 

\textbf{Parameters.} Our sparse density lists and trees have priors on the model structure, such as the number of leaves (Method I), branches (Method II), or length of the list (Method III). In Table \ref{tableS:params}, we summarize all parameters that one needs to define in order to run our methods. Pseudocounts are typically set to small values in order to avoid zero densities. For all methods, $\lambda$ regularizes the complexity of the model and reflects the prior belief on how sparse the user expects/would like the model to be. 
However, the resulting model complexity also depends on the data distribution.
To give specific examples of priors and optimal model complexities, we analyzed trees and lists that we computed while evaluating run time in Appendix \ref{appendix:runtime}. For every dataset, we reported the prior value that led to the maximum log-likelihood model and the complexity of this model (see Table \ref{tableS:best_params}). For example, for the COMPAS dataset with $\sim6500$ training samples and $19$ feature-value pairs, a prior of $8$ on the number of leaves led to a tree with $22$ leaves; a prior of $2$ on the number of branches led to a tree with $32$ leaves; a prior of $3$ on the length of the rule list led to a model of length $5$.
While Table \ref{tableS:best_params} is a posthoc analysis of experiments conducted in Appendix \ref{appendix:runtime}, it can still serve as a reference point for the prior value and optimal model complexity for different datasets. We also encourage users to perform cross-validation to choose parameters similar to the experiments we conducted in Section \ref{sec:experiments} and Appendix \ref{appendix:runtime}.

\begin{table}[ht]
    \centering
    \begin{tabular}{p{0.05\textwidth}|p{0.3\textwidth}|p{0.5\textwidth}}
    \hline 
    P & Meaning & Recommendations \\\hline \addlinespace  
    \multicolumn{3}{c}{Leaf-Sparse Density tree}\\ \hline
     $\lambda$ & Desired number of leaves in the tree.  & 8, 10. \newline Experimentally we found that setting the prior for the number of leaves to 8 achieves good cross-validation log-likelihood.\\\hline
      $\alpha$   & Pseudocount, used to avoid zero values for the estimated densities. & 1, 2 \newline A small value.\\\hline\addlinespace \addlinespace

      \multicolumn{3}{c}{Branch-Sparse Density tree}\\\hline
      $ \lambda$ & Desired number of branches at each internal node of the tree. & 2, 3 \newline Depends on the number of categories for each feature, and how much the user would like to aggregate them. If sparsity is preferred, then 2 or 3 is a good choice. Otherwise, may be set to 4 or 5. \\\hline
      $\alpha$   & Pseudocount, used to avoid zero values for the estimated densities. & 1, 2 \newline A small value.\\\hline\addlinespace \addlinespace

\multicolumn{3}{c}{Sparse Density Rule List}\\\hline
      $ \lambda$ & Desired length of the rule list. & 7 or higher (for larger datasets). \\\hline
      $\eta$ & Desired number of conjunctions in a rule. &  1, 2, or 3 (for a smaller number of feature/value pairs). \newline For larger datasets, mining rules and storing the data can be memory-expensive, so smaller values of this parameter may be preferred. \\\hline
      $\alpha$   & Pseudocount, used to avoid zero values for the estimated densities. & 1, 2 \newline A small value.\\\hline
      
    \end{tabular}
    \caption{Description of the parameters for sparse density trees and lists. Additionally, initial guidance regarding the values of parameters is provided.}
    \label{tableS:params}
\end{table}

\begin{table}[ht]
\centering
\small
\begin{tabular}{|p{0.17\textwidth}|p{0.07\linewidth}|p{0.07\textwidth}||p{0.075\textwidth}|p{0.085\textwidth}||p{0.075\textwidth}|p{0.085\textwidth}||p{0.075\textwidth}|p{0.075\textwidth}|p{0.085\textwidth}|}
\hline 
\multicolumn{3}{|c|}{Data set} & \multicolumn{2}{c}{Leaf-based} &\multicolumn{2}{|c|}{Branch-based} &\multicolumn{3}{c|}{Rule list} \\\hline \addlinespace \hline
	&	 Number of train points 	&	 Number of feature-value pairs 	&	Prior, $\lambda$	&	Number of leaves in the optimal model	&	Prior, $\lambda$	&	Number of leaves in the optimal model	&	Prior, $\lambda$	&	Prior, $\eta$	&	Length of the optimal model	\\\hline
        Balance  	&	531	&	20	&	5	&	2	&	3	&	40	&	3	&	1	&	3	\\\hline
        Bank Full 	&	38429	&	51	&	8	&	100	&	2	&	52	&	5	&	2	&	8	\\\hline
        Car 	&	1468	&	21	&	8	&	2	&	3	&	31	&	3	&	1	&	4	\\\hline
        Chess (King-Rook vs. King-Pawn) 	&	2716	&	73	&	5	&	26	&	3	&	47	&	7	&	2	&	8	\\\hline
        COCO stuff+thing labels hierarchy 	&	1607459	&	206	&	5	&	202	&	2	&	208	&	3	&	2	&	6	\\\hline
        COCO stuff labels 	&	118280	&	182	&	8	&	32	&	2	&	41	&	7	&	2	&	9	\\\hline
        COCO thing labels 	&	117266	&	160	&	5	&	47	&	3	&	29	&	7	&	1	&	7	\\\hline
        COMPAS 	&	6489	&	19	&	8	&	22	&	2	&	32	&	3	&	1	&	5	\\\hline
        Connect 4 	&	57423	&	126	&	10	&	47	&	3	&	42	&	7	&	2	&	7	\\\hline
        Crime 	&	3178	&	24	&	5	&	20	&	2	&	40	&	7	&	1	&	7	\\\hline
        HELOC 	&	8890	&	195	&	8	&	84	&	3	&	93	&	7	&	1	&	8	\\\hline
        Mushroom 	&	4797	&	100	&	10	&	52	&	2	&	73	&	3	&	1	&	6	\\\hline
        Nursery 	&	11016	&	27	&	8	&	2	&	3	&	48	&	3	&	2	&	2	\\\hline
        Telco Customer Churn  	&	5977	&	73	&	8	&	41	&	3	&	73	&	7	&	1	&	10	\\\hline
        Titanic 	&	1761	&	8	&	5	&	11	&	2	&	13	&	5	&	1	&	7	\\\hline
        US Census 1990 (1m) 	&	1000000	&	396	&	5	&	78	&	3	&	82	&	7	&	1	&	7	\\\hline
        US Census 1990 	&	2089542	&	396	&	5	&	100	&	2	&	105	&	7	&	1	&	5	\\\hline
\end{tabular}
\caption{Description of priors and the model complexities for models that maximized log-likelihood during the tuning procedure described in Appendix \ref{appendix:runtime}.}\label{tableS:best_params}
\end{table}



\end{APPENDICES}
\end{document}